\documentclass[11pt]{article}
\usepackage{geometry}
\usepackage{cite}
\usepackage{amsmath,amssymb,amsthm,amsfonts}
\usepackage{mathtools}
\usepackage{algorithm}
\usepackage{algpseudocode}
\usepackage{algpseudocodex}
\usepackage{graphicx}
\usepackage{textcomp}
\usepackage{xcolor}
\usepackage{hyperref}
\usepackage[capitalise]{cleveref}
\usepackage{caption}
\usepackage{svg}
\usepackage{subcaption}
\usepackage{booktabs}
\usepackage{multirow}
\usepackage{tikz}
\usetikzlibrary{shapes.geometric}
\definecolor{mppiblue}{HTML}{5B9BFF}
\definecolor{ramppipurple}{HTML}{C56BE3}
\definecolor{drmppigreen}{HTML}{55B8AC}
\definecolor{dramppiorange}{HTML}{FFAA5C}

\definecolor{obstaclepink}{HTML}{EB1D87}
\definecolor{obstaclegreen}{HTML}{34980C}
\definecolor{obstaclered}{HTML}{FF0073}

\newcommand{\legendbox}[2]{%
    \textcolor{#1}{\rule{1.2ex}{1.2ex}}\hspace{0.3em}#2%
}

\newcommand{\legendstar}[1]{%
    \tikz[baseline=-0.6ex]
    \node[
        star,
        star points=5,
        star point ratio=2.25,
        fill=red,
        draw=red,
        minimum size=1.2ex,
        inner sep=0pt
    ] {};
    \hspace{-0.05em}#1%
}
\crefformat{equation}{(#2#1#3)}
\crefrangeformat{equation}{(#3#1#4)--(#5#2#6)}
\crefmultiformat{equation}
  {(#2#1#3)}
  { and (#2#1#3)}
  {, (#2#1#3)}
  {, and (#2#1#3)}
  
\def\BibTeX{{\rm B\kern-.05em{\sc i\kern-.025em b}\kern-.08em
    T\kern-.1667em\lower.7ex\hbox{E}\kern-.125emX}}

\newcommand{\real}[1]{\mathbb{R}^{#1}}

\newcommand{\cvar}{\mathrm{CVaR}}

\newtheorem{lemma}{Lemma}
\newtheorem{example}{Example}
\newtheorem{remark}{Remark}
\newtheorem{definition}{Definition}
\newtheorem{proposition}{Proposition}
\newtheorem{corollary}{Corollary}

\title{DR$^2$-MPPI: Distributionally Robust Risk-Aware Model Predictive Path Integral Control%
\thanks{This work was supported by the Higher Education and Science Committee of the RA (Research Project \textnumero\,24FP-2C017) and by the Center for Scientific Innovation and Education.}
} %
\author{
Anna Beklaryan, Lusine Barseghyan, Haykanush Darbinyan, and
Astghik Hakobyan%
\thanks{The first two authors contributed equally.}
\thanks{A. Beklaryan, L. Barseghyan, and H. Darbinyan are with the National Polytechnic University of Armenia, Yerevan, Armenia.
A. Beklaryan is also with Yerevan State University, Yerevan, Armenia.
A. Hakobyan is with the National Polytechnic University of Armenia and American University of Armenia, Yerevan, Armenia (e-mail: {\tt\small astghik.hakobyan@aua.am}).}
}

\date{}

\begin{document}

\maketitle


\begin{abstract}
Model predictive path integral (MPPI) control provides efficient, parallelizable planning for nonlinear robotic systems, but risk-aware variants can underestimate danger when predicted obstacle distributions are inaccurate. We introduce distributionally robust risk-aware MPPI (DR$^2$-MPPI), which upper-bounds the worst-case conditional value-at-risk (CVaR) of signed-clearance loss over obstacle- and horizon-dependent Wasserstein ambiguity sets centered on empirical pose distributions. We prove that the signed-clearance margin loss is $1$-Lipschitz under an obstacle-specific pose metric and combine this result with the Wasserstein stability of CVaR to obtain a tractable bound: empirical CVaR plus an analytic robustness correction. This correction maps predictive-distribution mismatch directly to clearance inflation without auxiliary online optimization or changes to the sampling structure of MPPI. Across 100 runs per controller and simulation scenario, DR$^2$-MPPI produced no ground-vehicle collision runs and one Crazyflie collision run. In five hardware runs, it was the only evaluated configuration with no collisions and positive mean run-minimum signed clearance.
\end{abstract}


\section{Introduction}

Robots navigating in dynamic environments rely on predictions of how surrounding agents will move. Simplified models, noisy or intermittent observations, stochastic behavior, and finite samples can cause the predicted obstacle-motion distribution to differ from that encountered during execution. Treating this distribution as exact may therefore underestimate the risk of trajectories passing near obstacles.

Model predictive control (MPC) supports planning with uncertain predictions by incorporating system dynamics, constraints, and anticipated environmental changes into a receding-horizon problem~\cite{mayne2014model}. In safety-critical robotics, MPC has been applied to collision avoidance and probabilistic planning under uncertain obstacle motion~\cite{brito2019model,zhu2019chance,de2021scenario}. However, nonlinear MPC with nonconvex collision-avoidance objectives can be computationally demanding. Model predictive path integral (MPPI) control instead updates a nominal sequence from importance-weighted sampled trajectories~\cite{williams2018information}. Its derivative-free, parallelizable structure is attractive for real-time robotic control~\cite{enrico2025comparison}.

Recent MPPI extensions consider chance constraints, uncertain dynamic obstacles, external disturbances, model mismatch, and uncertain system parameters~\cite{gandhi2021robust,yin2024chance,mohamed2025chance,yoon2025uncertainty,zhang2026pred,vahs2026parameter,yin2023shield}. Risk-aware MPPI (RA-MPPI) uses conditional value-at-risk (CVaR) to penalize adverse outcomes in the tail of a loss distribution~\cite{rockafellar2000optimization,yin2023risk}. Dynamic risk-aware MPPI (DRA-MPPI) instead estimates collision probabilities from stochastic predictions of moving agents~\cite{trevisan2025dynamic}. Distributionally robust path-integral control has also been proposed for uncertainty in the system disturbance distribution~\cite{park2024distributionally}. In contrast, we consider distributional ambiguity in stochastic predictions of dynamic obstacle poses. Existing risk-aware MPPI methods that operate on such predictions remain tied to the reference distributions that generate their predictive samples~\cite{yin2023risk,trevisan2025dynamic}.

Distributionally robust MPC (DR-MPC) addresses such ambiguity by evaluating decisions over a family of plausible distributions, often defined by a Wasserstein neighborhood of a reference distribution~\cite{lu2020soft,coppens2021data,micheli2022data,mcallister2024distributionally}. Wasserstein robustness has also been studied in robotic planning, collision avoidance, and multi-robot navigation~\cite{lathrop2021distributionally,zolanvari2023wasserstein,hakobyan2022distributionally,hakobyan2021wasserstein,navsalkar2023data,safaoui2024distributionally}. These methods commonly rely on constrained optimization or safety-filter reformulations. Evaluating worst-case risk directly within MPPI would introduce auxiliary distributional optimization across many rollouts and prediction steps. This motivates our central question: \emph{Can distributional robustness be incorporated into MPPI without sacrificing its computational efficiency?}

We answer this question with distributionally robust risk-aware MPPI (DR$^2$-MPPI). For each obstacle and prediction step, we define risk as the worst-case CVaR of a signed-clearance margin loss over a $p$-Wasserstein ambiguity set centered on an empirical distribution of predicted obstacle poses. Combining the $p$-Wasserstein stability of CVaR with the Lipschitz continuity of the loss yields a tractable upper bound consisting of the empirical CVaR and an analytic correction. This correction converts distributional ambiguity into obstacle- and horizon-dependent clearance inflation while preserving MPPI's sampling and control-update structure. Our contributions are:

\begin{itemize}
\item We formulate a distributionally robust obstacle-risk measure for MPPI that accounts for stochastic obstacle motion and ambiguity in the predictive distribution.

\item We prove that the signed-clearance margin loss is $1$-Lipschitz under an obstacle-specific pose metric accounting for translation, rotation, and obstacle size. The resulting bound maps ambiguity radii to clearance margins without requiring online distributional optimization.

\item We compare DR$^2$-MPPI with MPPI, RA-MPPI, and DRA-MPPI in ground-vehicle and Crazyflie simulations. Across $100$ runs per controller and scenario, DR$^2$-MPPI records no ground-vehicle collision runs and one Crazyflie collision run, with mean planning times comparable to MPPI and RA-MPPI.

\item We demonstrate real-world feasibility in Crazyflie hardware experiments comprising five runs per controller configuration, where DR$^2$-MPPI is the only evaluated configuration with no collision runs and positive mean run-minimum signed clearance.
\end{itemize}

\section{Preliminaries}

\subsection{Robot and Obstacle Modeling}

We consider a discrete nonlinear dynamical system
\begin{equation*}
    x(t+1)=f(x(t),v(t)),
\end{equation*}
where $x(t)\in\real{n}$ denotes the robot state and $f:\real{n}\times\real{m}\to\real{n}$ represents the system dynamics. In MPPI, candidate rollout controls are sampled as $v(t)=u(t)+\tilde{u}(t)$, where $u(t)\in\real{m}$ is the nominal control and $\tilde{u}(t)\sim\mathcal{N}(0,\Sigma)$, with $\Sigma\in\mathbb{S}_{++}^m$. These perturbations generate the rollout ensemble used to update the nominal control sequence.

The robot operates in a $d$-dimensional workspace containing $n_o$ obstacles, each of which may be static or dynamic. For every admissible state $x$, let $\mathcal{B}(x)\subset\mathbb{R}^d$ denote the nonempty, compact, and convex region occupied by the robot. This footprint may depend on the position and orientation components of $x$. For obstacle $i$, let $\mathcal{O}_i^0\subset\mathbb{R}^d$ be a nonempty, compact, and convex body-frame representation of its occupied region. Both robot and obstacle geometries may be represented, or conservatively approximated, by polytopes, ellipsoids, spheres, or solid cylinders.\footnote{For example, a generic bounded polytope may be written as
$\mathcal{S}=\{z\in\mathbb{R}^d\mid Hz\leq h\}$, while a centered ellipsoid may be written as $\mathcal{S}=\{z\in\mathbb{R}^d\mid z^\top Wz\leq1\}$, with $W\succ0$. These representations may be used for either $\mathcal{B}(x)$ or $\mathcal{O}_i^0$, after applying the appropriate pose transformation.}

Let $\theta_i=(p_i,R_i)\in\Xi\coloneqq \mathbb{R}^d\times SO(d)$ denote the obstacle pose, where $p_i\in\mathbb{R}^d$ and $R_i\in SO(d)$ are its position and orientation. Its occupied region in the workspace is
\begin{equation*}
\mathcal{O}_i(\theta_i)
\coloneqq
R_i\mathcal{O}_i^0+p_i.
\end{equation*}
At time $t$, the obstacle pose is $\theta_{i,t}=(p_{i,t},R_{i,t})$, and we write $\mathcal{O}_i(t) \coloneqq \mathcal{O}_i(\theta_{i,t})$.

The future pose $\theta_{i,t+k}$ is uncertain because of perception and prediction errors, stochastic obstacle motion, and unmodeled interactions. At planning time $t$, it is modeled as a random variable with an unknown predictive distribution $\mathbb{P}_{\theta,k\mid t}^i$. The planner has access only to an estimated reference distribution $\hat{\mathbb{P}}_{\theta,k\mid t}^i$, obtained from a perception or motion-prediction module and represented empirically by predicted pose samples. No particular parametric obstacle-motion model is required. The possible mismatch between $\mathbb{P}_{\theta,k\mid t}^i$ and $\hat{\mathbb{P}}_{\theta,k\mid t}^i$ motivates the distributionally robust formulation developed subsequently.

\begin{example}\label{ex:pract_ex}
An online tracking or prediction module may approximate the pose evolution of obstacle $i$ as
\begin{equation}\label{eqn:obs_kin}
\begin{aligned}
    p_{i,t+1}
    &=
    p_{i,t}
    +\hat{v}_{i,t}\Delta t
    +\tilde{p}_{i,t},\\
    R_{i,t+1}
    &=
    R_{i,t}
    \exp\!\left(
        (\hat{\omega}_{i,t}\Delta t
        +\tilde{\psi}_{i,t})^\wedge
    \right),
\end{aligned}
\end{equation}
where $\hat{v}_{i,t}$ and $\hat{\omega}_{i,t}$ are the estimated linear and angular velocities, while $\tilde{p}_{i,t}$ and $\tilde{\psi}_{i,t}$ represent stochastic prediction errors. Samples generated by such a model can be used to construct $\hat{\mathbb{P}}_{\theta,k\mid t}^i$. Equation~\eqref{eqn:obs_kin} is illustrative; the proposed framework does not depend on this particular model.
\end{example}

Given a prescribed clearance $d_s\geq0$, the nominal collision-avoidance condition is
\begin{equation}\label{eq:safety_distance}
\operatorname{sdist}\!\left(
\mathcal{B}(x(t)),
\mathcal{O}_i(\theta_{i,t})
\right)
\geq d_s,
\qquad
i=1,\ldots,n_o,
\end{equation}
where $\operatorname{sdist}$ denotes signed clearance between the robot and obstacle regions. It is positive when their interiors are separated, zero at contact, and negative when they overlap. Thus, $d_s=0$ places the nominal safety boundary at physical contact.\footnote{For nonempty, compact, convex sets $\mathcal{A},\mathcal{C}\subset\mathbb{R}^d$, we define $\operatorname{sdist}(\mathcal{A},\mathcal{C}) \coloneqq \max_{\|n\|_2=1} \left( \inf_{c\in\mathcal{C}} n^\top c
- \sup_{a\in\mathcal{A}} n^\top a \right)$. When the sets overlap, the magnitude of the negative value is the minimum translational separation required to eliminate
the penetration.}

Because the executed robot trajectory and future obstacle poses are uncertain, requiring~\eqref{eq:safety_distance} to hold for every realization is generally impractical. We instead quantify signed-clearance violations using risk-sensitive measures and account for the horizon-dependent ambiguity in the predicted obstacle distribution.

\subsection{MPPI Overview}

MPPI is a derivative-free sampling method for nonlinear stochastic optimal control~\cite{williams2018information}. At each planning step, it propagates perturbed control sequences through the dynamics and updates a nominal sequence using exponentially weighted trajectory costs.

Consider a nominal control sequence $\mathbf{u} \coloneqq (u_0,\ldots,u_{K-1})$, where $K$ denotes the prediction horizon. MPPI generates sampled control sequences according to 
\[
v_k = u_k+\tilde{u}_k, \quad \tilde{u}_k\sim\mathcal{N}(0,\Sigma),
\]
for $k=0,\ldots,K-1$, where $\Sigma\in\mathbb{S}_{++}^{m}$ denotes the control-sampling covariance. Each sampled sequence $\mathbf{v}=(v_0,\ldots,v_{K-1})$ induces a state trajectory through $ x_{k+1}=f(x_k,v_k)$, $x_0=x(t)$.

Given a state cost $c_x:\real{n}\rightarrow\mathbb{R}$, a terminal cost $c_K:\real{n}\rightarrow\mathbb{R}$, and a positive semidefinite control-cost matrix $R_u\succeq0$, we define the running task cost as
\begin{equation}\label{eqn:mppi_running_cost}
    c(x_k,v_k)
    \coloneqq
    c_x(x_k)
    +
    \frac{1}{2}v_k^\top R_u v_k.
\end{equation}
The corresponding trajectory cost of the sampled control sequence is
\begin{equation}\label{eqn:mppi_state_cost}
    S(\mathbf{v})
    \coloneqq
    c_K(x_K)
    +
    \sum_{k=0}^{K-1}
    c(x_k,v_k).
\end{equation}
The quadratic term in~\cref{eqn:mppi_running_cost} is an application-dependent control-effort penalty. For $M$ sampled rollouts, let $S^m \coloneqq S(\mathbf{v}^m)$. Accounting for the change of measure between the nominal and sampled Gaussian control distributions gives the normalized weights
\begin{equation}\label{eqn:mppi_weights}
    w^m
    =
    \frac{
        \exp\!\left[
            -\frac{1}{\lambda}
            \left(
                S^m
                +
                \lambda
                \sum_{k=0}^{K-1}
                u_k^\top\Sigma^{-1}v_k^m
            \right)
        \right]
    }{
        \displaystyle
        \sum_{j=1}^{M}
        \exp\!\left[
            -\frac{1}{\lambda}
            \left(
                S^j
                +
                \lambda
                \sum_{k=0}^{K-1}
                u_k^\top\Sigma^{-1}v_k^j
            \right)
        \right]
    }.
\end{equation}
A common exponent offset may be used for numerical stability. The control update is
\begin{equation}\label{eqn:mppi_update}
    u_k^\star
    \approx
    \sum_{m=1}^{M}w^m v_k^m
    =
    u_k
    +
    \sum_{m=1}^{M}w^m\tilde{u}_k^m,
    \;
    k=0,\ldots,K-1.
\end{equation}

Only the first updated control is applied before the horizon is shifted. The same rollout ensemble provides the candidate trajectories on which obstacle risk is evaluated.

\subsection{Nominal Risk-Aware Construction}

DR$^2$-MPPI builds on a nominal risk-aware cost that augments each sampled rollout with a measure of obstacle risk~\cite{yin2023risk}. We quantify this risk using CVaR.

Given a scalar loss $\ell(\xi)$ associated with a random variable $\xi\sim\mathbb{P}_{\xi}$, the CVaR at confidence level $\alpha\in(0,1)$ is defined as~\cite{rockafellar2000optimization,rockafellar2002conditional}
\begin{equation*}
\cvar_{\alpha}^{\mathbb{P}_{\xi}}
\left(\ell(\xi)\right)
\coloneqq
\inf_{z\in\mathbb{R}}
\left\{
z+
\frac{1}{1-\alpha}
\mathbb{E}_{\mathbb{P}_{\xi}}
\left[
\left(\ell(\xi)-z\right)^+
\right]
\right\},
\end{equation*}
where $(a)^+\coloneqq\max\{a,0\}$. CVaR measures losses in the upper $(1-\alpha)$ tail, accounting for both the occurrence and magnitude of adverse outcomes.

We define the signed-clearance margin loss for robot state $x$ and obstacle pose $\theta_i$ as
\begin{equation}\label{eqn:loss}
\ell(x,\theta_i)
\coloneqq
d_s-
\operatorname{sdist}\!\left(
\mathcal{B}(x),
\mathcal{O}_i(\theta_i)
\right).
\end{equation}
The loss is positive when the prescribed clearance is violated, zero at the clearance boundary, and negative otherwise. Unlike the unsigned-distance loss, it continues to increase after the robot and obstacle regions overlap.

At planning time $t$, the nominal risk associated with obstacle $i$, prediction step $k$, and candidate robot state $x_k$ is
\begin{equation}\label{eqn:risk_measure}
\rho_{i,k|t}(x_k)
\coloneqq
\cvar_{\alpha}^{
\hat{\mathbb{P}}_{\theta,k|t}^{i}
}
\left(
\ell(x_k,\theta)
\right),
\end{equation}
where $\theta$ denotes the random obstacle pose governed by $\hat{\mathbb{P}}_{\theta,k|t}^{i}$.

For the $m$-th MPPI rollout, the obstacle-wise risks are aggregated as
\begin{equation*}
\rho_{k|t}^{m}
\coloneqq
\max_{i=1,\ldots,n_o}
\rho_{i,k|t}(x_k^m).
\end{equation*}

Following the risk-aware MPPI construction, the nominal trajectory cost $S^m$ in~\cref{eqn:mppi_state_cost} is augmented whenever the risk exceeds a prescribed threshold $\bar{\rho}$:
\begin{equation*}
\widetilde{S}^m
\coloneqq
S^m+
\beta
\sum_{k=0}^{K-1}
\left[
\rho_{k|t}^m-\bar{\rho}
\right]^+,
\end{equation*}
where $\beta>0$ is the risk-penalty weight. Rollouts with larger risk violations consequently receive smaller importance weights.

The risk-aware weights are obtained from~\eqref{eqn:mppi_weights} by replacing $S^m$ with $\widetilde{S}^m$. The nominal control sequence is then updated using~\eqref{eqn:mppi_update}. Thus, risk awareness modifies the rollout cost while preserving the MPPI sampling and control-update mechanism.

\begin{remark}\label{rem:uncertainty_decomp}
The risk in~\cref{eqn:risk_measure} quantifies obstacle-prediction uncertainty for a fixed candidate robot state. Robot-trajectory variation is represented by the MPPI
rollout samples, whereas distributional ambiguity is applied only to the predicted obstacle pose. A joint-risk formulation over robot and obstacle uncertainty could instead be constructed using joint samples, at the cost of additional computation.
\end{remark}

\section{Distributionally Robust Risk-Aware MPPI}

The preceding formulation evaluates obstacle risk under a reference predictive distribution estimated from finite, noisy data. Mismatch with the true obstacle-motion distribution may therefore cause CVaR to underestimate tail risk.

We address this mismatch by replacing nominal CVaR with its worst-case value over a Wasserstein ambiguity set centered on the reference distribution. This yields a tractable distributionally robust risk measure that integrates into the MPPI rollout cost without altering the sampling or control-update mechanism.

\subsection{Distributionally Robust Safety Risk}

We begin by recalling the Wasserstein distance used to quantify discrepancies between probability distributions~\cite{villani2009optimal}. For $p\geq1$, let $\mathcal{P}_p(\Xi)$ denote the set of Borel probability measures on $\Xi$ with finite $p$-th moment.\footnote{That is, for some $\xi_0\in\Xi$, $\int_{\Xi}d_\Xi(\xi,\xi_0)^p\,\mathbb{P}(d\xi)<\infty$.}

\begin{definition}[Wasserstein distance]
Let $(\Xi,d_\Xi)$ be a metric space. For $p\geq1$, the $p$-Wasserstein distance between $\mathbb{P},\mathbb{Q}\in\mathcal{P}_p(\Xi)$ is
\begin{equation*}
W_p(\mathbb{P},\mathbb{Q})
\coloneqq
\left(
\inf_{\pi\in\Pi(\mathbb{P},\mathbb{Q})}
\int_{\Xi\times\Xi}
d_\Xi(\xi,\xi')^p
\,\pi(d\xi,d\xi')
\right)^{1/p},
\end{equation*}
where $\Pi(\mathbb{P},\mathbb{Q})$ denotes the set of all
couplings of $\mathbb{P}$ and $\mathbb{Q}$.
\end{definition}

For obstacle $i$, define its body-frame radius as $r_i\coloneqq \sup_{z\in\mathcal{O}_i^0}\|z\|_2$, which is finite by compactness and positive for the nondegenerate obstacle geometries considered here. We equip the common pose space $\Xi=\mathbb{R}^d\times SO(d)$ with the obstacle-specific ground metric
\begin{equation}\label{eqn:pose_metric}
d_{\Xi_i}(\theta,\theta')
\coloneqq
\|p-p'\|_2
+
r_i\|R-R'\|_{\mathrm{op}},
\end{equation}
where $\theta=(p,R)$, $\theta'=(p',R')$, and $\|\cdot\|_{\mathrm{op}}$ denotes the induced operator norm. The translation and rotation terms have consistent physical units, while the rotational contribution reflects the size of obstacle $i$. Let $W_{p,i}$ denote the $p$-Wasserstein distance induced by $d_{\Xi_i}$.

At planning time $t$, let $\{\theta_{i,t+k|t}^j\}_{j=1}^{N}$ denote $N$ predicted pose samples for obstacle $i$ at prediction step $k$. We represent the corresponding reference predictive distribution by the empirical measure
\begin{equation}\label{eqn:empirical_obs_dist}
\hat{\mathbb{P}}_{\theta,k|t}^i
\coloneqq
\frac{1}{N}
\sum_{j=1}^{N}
\delta_{\theta_{i,t+k|t}^j},
\end{equation}
where $\delta_\theta$ denotes the Dirac measure concentrated at $\theta$. Since~\cref{eqn:empirical_obs_dist} has finite support, $\hat{\mathbb{P}}_{\theta,k|t}^i\in\mathcal{P}_p(\Xi)$ for every $p\geq1$.

Following standard data-driven Wasserstein distributionally robust optimization constructions~\cite{mohajerin2018data,gao2023distributionally}, we assign a radius $\varepsilon_{i,k|t}\geq0$ to obstacle $i$ at prediction step $k$ and define
\begin{equation*}
\mathbb{D}_{\varepsilon_{i,k|t},p}^{\,i}
\left(
\hat{\mathbb{P}}_{\theta,k|t}^{i}
\right)
\coloneqq
\left\{
\mathbb{P}\in\mathcal{P}_p(\Xi)
\;\middle|\;
W_{p,i}\left(
\mathbb{P},
\hat{\mathbb{P}}_{\theta,k|t}^{i}
\right)
\leq\varepsilon_{i,k|t}
\right\}.
\end{equation*}
The radius $\varepsilon_{i,k|t}$ specifies the magnitude of distributional mismatch considered for that obstacle and prediction step. This allows the ambiguity set to reflect differences in prediction reliability across obstacles and along the horizon. Obstacles with a common prediction model and comparable error statistics may share the same horizon-dependent radius sequence. Setting $\varepsilon_{i,k|t}=0$ recovers the corresponding reference predictive distribution.

Motivated by Wasserstein DR-CVaR formulations for collision avoidance~\cite{hakobyan2021wasserstein,navsalkar2023data}, we define the distributionally robust risk associated with obstacle $i$, prediction step $k$, and candidate robot state $x_k$ as
\begin{equation}\label{eqn:drcvar}
\rho_{i,k|t}^{\mathrm{DR},p}(x_k)
\coloneqq
\sup_{\mathbb{P}\in
\mathbb{D}_{\varepsilon_{i,k|t},p}^{\,i}
\left(
\hat{\mathbb{P}}_{\theta,k|t}^{i}
\right)
}
\cvar_{\alpha}^{\mathbb{P}}
\left(
\ell(x_k,\theta)
\right).
\end{equation}
This quantity is the largest CVaR of the signed-clearance margin loss over the obstacle-pose distributions admitted by the ambiguity set. Unlike the nominal risk measure in~\eqref{eqn:risk_measure}, it accounts for deviations from the reference predictive distribution.

Now, let $\mathbb{P}_{\theta,k|t}^i$ denote the underlying predictive distribution of obstacle $i$. If
$\mathbb{P}_{\theta,k|t}^i
\in
\mathbb{D}_{\varepsilon_{i,k|t},p}^{\,i}
\bigl(
\hat{\mathbb{P}}_{\theta,k|t}^i
\bigr)$, then
$\cvar_{\alpha}^{\mathbb{P}_{\theta,k|t}^i}
\left(
\ell(x_k,\theta)
\right)
\leq
\rho_{i,k|t}^{\mathrm{DR},p}(x_k)$.
This guarantee is conditional on ambiguity-set membership and therefore depends on the calibration of $\varepsilon_{i,k|t}$.

\begin{remark}[Radius selection]
\label{rem:radius_selection}
Under suitable moment assumptions, Wasserstein concentration bounds can determine a sampling radius that contains the distribution generating the i.i.d.\ prediction samples with prescribed confidence~\cite{fournier2015rate,mohajerin2018data}. Prediction-model bias can be incorporated through an additional radius calibrated from held-out prediction errors, yielding
$\varepsilon_{i,k|t}
=
\varepsilon_{i,k|t}^{\mathrm{samp}}
+
\varepsilon_{i,k|t}^{\mathrm{mod}}$.
The fixed radii used in our experiments are selected empirically and are not claimed to provide statistical coverage guarantees.
\end{remark}

\subsection{Tractable DR-CVaR Bound}

Direct evaluation of~\cref{eqn:drcvar} requires optimization over an infinite-dimensional set of probability measures and is therefore unsuitable for evaluation across MPPI rollouts. We instead use the stability of CVaR with respect to the $p$-Wasserstein distance. The following stability bound follows from the dual representation of CVaR~\cite{shapiro2012minimax} and H\"older's inequality.

\begin{proposition}[Wasserstein stability of CVaR]
\label{prop:drcvar_reform}
Let $(\Xi,d_\Xi)$ be a metric space, let $p\geq1$, and let $\mathbb{P},\mathbb{Q}\in\mathcal{P}_p(\Xi)$. If $g:\Xi\to\mathbb{R}$ is Lipschitz continuous with constant
$\mathrm{Lip}(g)$, then, for every $\alpha\in(0,1)$,
\begin{equation}\label{eqn:cvar_stability}
\left|
\cvar_{\alpha}^{\mathbb{P}}\!\left(g(\xi)\right)
-
\cvar_{\alpha}^{\mathbb{Q}}\!\left(g(\xi)\right)
\right|
\leq
\frac{\mathrm{Lip}(g)}
{(1-\alpha)^{1/p}}
W_p(\mathbb{P},\mathbb{Q}).
\end{equation}
Consequently, for the $p$-Wasserstein ball $\mathbb{D}_{\eta,p}(\hat{\mathbb{P}})$ of radius $\eta\geq0$ centered at $\hat{\mathbb{P}}\in\mathcal{P}_p(\Xi)$,
\begin{equation}\label{eqn:drcvar_ineq}
\sup_{\mathbb{P}\in
\mathbb{D}_{\eta,p}(\hat{\mathbb{P}})}
\cvar_{\alpha}^{\mathbb{P}}
\left(g(\xi)\right)
\leq
\cvar_{\alpha}^{\hat{\mathbb{P}}}
\left(g(\xi)\right)
+
\frac{\eta}{(1-\alpha)^{1/p}}
\mathrm{Lip}(g).
\end{equation}
\end{proposition}

The proof is provided in Appendix~\ref{app:drcvar_reform}. We next specialize this stability result to the signed-clearance margin loss.

\begin{lemma}\label{lem:lipschitz_loss}
Suppose that the robot and obstacle regions are nonempty, compact, and convex. For any fixed robot state $x$ and obstacle $i\in\{1,\ldots,n_o\}$, the signed-clearance margin loss $\theta\mapsto\ell(x,\theta)$ defined in~\cref{eqn:loss} is $1$-Lipschitz with respect to $d_{\Xi_i}$; that is,
\begin{equation*}
\left|
\ell(x,\theta)-\ell(x,\theta')
\right|
\leq
d_{\Xi_i}(\theta,\theta'),
\qquad
\forall\,\theta,\theta'\in\Xi.
\end{equation*}
\end{lemma}

The proof is given in Appendix~\ref{app:lipschitz_loss}. The robot state, and hence its footprint $\mathcal{B}(x)$, is fixed in~\cref{lem:lipschitz_loss}. Robot-trajectory variation is represented by the MPPI rollout samples, whereas distributional ambiguity is applied only to the predicted obstacle pose.

Combining~\cref{prop:drcvar_reform,lem:lipschitz_loss} gives the following tractable bound.

\begin{corollary}[Tractable DR-CVaR safety risk]
\label{cor:safety_drcvar}
Consider the pose space $\Xi$ equipped with the obstacle-specific metric $d_{\Xi_i}$ in~\eqref{eqn:pose_metric}. For obstacle $i$, prediction step $k$, and candidate robot state $x_k$, let $\hat{\mathbb{P}}_{\theta,k|t}^{i}$ be the empirical predictive distribution in~\cref{eqn:empirical_obs_dist}, and let $\varepsilon_{i,k|t}\geq0$ be its Wasserstein radius. Then, for every $p\geq1$ and $\alpha\in(0,1)$,
\begin{equation}\label{eqn:tract_drcvar}
\rho_{i,k|t}^{\mathrm{DR},p}(x_k)
\leq
\rho_{i,k|t}(x_k)
+
\frac{\varepsilon_{i,k|t}}
{(1-\alpha)^{1/p}}.
\end{equation}
\end{corollary}

\begin{proof}
Apply~\cref{prop:drcvar_reform} with $g(\theta)=\ell(x_k,\theta)$, $\hat{\mathbb{P}}=\hat{\mathbb{P}}_{\theta,k|t}^{i}$, $\eta=\varepsilon_{i,k|t}$, and ground metric
$d_{\Xi_i}$. By~\cref{lem:lipschitz_loss}, $\mathrm{Lip}(g)\leq1$, and the empirical CVaR term equals $\rho_{i,k|t}(x_k)$ by~\cref{eqn:risk_measure}. The
left-hand side of~\eqref{eqn:drcvar_ineq} equals $\rho_{i,k|t}^{\mathrm{DR},p}(x_k)$ by~\cref{eqn:drcvar}, which proves the result.
\end{proof}

The bound in~\cref{eqn:tract_drcvar} replaces the infinite-dimensional worst-case risk evaluation with an empirical CVaR computation and the obstacle- and horizon-dependent correction $\varepsilon_{i,k|t}/(1-\alpha)^{1/p}$. It can therefore be evaluated within each MPPI rollout without solving an auxiliary distributional optimization problem.

For a fixed numerical radius $\eta\geq0$ and $p\geq1$, $W_{p,i}(\mathbb{P},\mathbb{Q})\geq W_{1,i}(\mathbb{P},\mathbb{Q})$ implies $\mathbb{D}_{\eta,p}^{\,i}(\hat{\mathbb{P}}) \subseteq \mathbb{D}_{\eta,1}^{\,i}(\hat{\mathbb{P}})$. Thus, $p=1$ gives the largest ambiguity set among these choices for a fixed radius. The bound in~\cref{eqn:tract_drcvar} also depends on $p$ through $(1-\alpha)^{-1/p}$. Comparisons across Wasserstein orders require separately calibrated radii because the same numerical radius defines different ambiguity sets. We use $p=1$ in the experiments, for which the correction reduces to $\varepsilon_{i,k|t}/(1-\alpha)$.

\begin{remark}[Systematic clearance inflation]
\label{rem:clearance_inflation}
For fixed $\alpha$ and $p$, the correction is independent of the MPPI rollout but may vary across obstacles and prediction steps. By the translation equivariance of CVaR and the definition of the loss in~\cref{eqn:loss}, adding the correction is equivalent to replacing $d_s$ with the effective clearance $d_{s,i,k|t}^{\mathrm{eff}}
=d_s+\varepsilon_{i,k|t}/(1-\alpha)^{1/p}$. When $d_s=0$, the effective clearance is supplied entirely by the Wasserstein correction. This expression maps each
Wasserstein radius, representing the admitted distributional mismatch, directly to a clearance inflation. When the radii are calibrated from prediction errors, the formulation therefore provides a systematic, distributionally grounded mechanism for determining obstacle- and horizon-dependent safety margins. The constant radii and $p=1$ used in our experiments are a special case.
\end{remark}

\subsection{DR$^2$-MPPI Algorithm}

The proposed DR$^2$-MPPI method is summarized in~\Cref{alg:drmppi}. At each planning iteration, the motion predictor provides $N$ pose samples for every obstacle and prediction step, defining the empirical distributions in~\eqref{eqn:empirical_obs_dist}. A Wasserstein radius $\varepsilon_{i,k|t}$ is assigned according to the prescribed radius schedule. The controller then samples $M$ perturbed control sequences and propagates the corresponding trajectories over the prediction horizon. For each rollout state, it evaluates the signed-clearance margin loss against the predicted obstacle-pose samples. The empirical CVaR is computed exactly from the ordered losses using~\Cref{alg:drcvar_eval}~\cite{rockafellar2002conditional}. The tractable upper bound is then $\bar{\rho}_{i,k|t}^{\mathrm{DR},m} = \hat{\rho}_{i,k|t}^{m} + \varepsilon_{i,k|t}/(1-\alpha)^{1/p}$.

At each prediction step, the maximum bound across obstacles defines the rollout safety risk. This risk augments the trajectory cost, after which the standard MPPI weighting and control update are applied. Thus, obstacle- and horizon-dependent distributional ambiguity is incorporated without solving an auxiliary distributional optimization problem inside each rollout.

\begin{algorithm}[t!]
\caption{DR$^2$-MPPI}
\label{alg:drmppi}
\begin{algorithmic}[1]
\State \textbf{Given:}
$\lambda,\Sigma,R_u,\alpha,
\{\varepsilon_{i,k|t}\}_{i,k},
\beta,\bar{\rho},K,M,N,p$

\State \textbf{Input:}
$\mathbf{u}=\{u_k\}_{k=0}^{K-1}$

\While{\textit{task not complete}}
    \State $x_0\gets x(t)$

    \For{$k\gets0$ to $K-1$}
        \For{$i\gets1$ to $n_o$}
            \State Obtain
            $\{\theta_{i,t+k|t}^{j}\}_{j=1}^{N}$
        \EndFor
    \EndFor

    \For{$m\gets1$ to $M$}
        \State Draw
        $\tilde{u}_k^m\sim\mathcal{N}(\mathbf{0},\Sigma)$,
        $k=0,\ldots,K-1$
        \State $x_0^m\gets x_0$, \quad $S^m\gets0$

        \For{$k\gets0$ to $K-1$}
            \State $v_k^m\gets u_k+\tilde{u}_k^m$
            \State $x_{k+1}^m\gets f(x_k^m,v_k^m)$
            \State $S^m\gets S^m+c(x_k^m,v_k^m)$
        \EndFor

        \State $S^m\gets S^m+c_K(x_K^m)$

        \For{$k\gets0$ to $K-1$}
            \For{$i\gets1$ to $n_o$}
                \For{$j\gets1$ to $N$}
                    \State
                    $L_{i,k}^{m,j}
                    \gets
                    \ell(x_k^m,\theta_{i,t+k|t}^{j})$
                \EndFor
                \State
                $\bar{\rho}_{i,k|t}^{\mathrm{DR},m}
                \gets
                \Call{EvalDRCVaR}{
                \{L_{i,k}^{m,j}\}_{j=1}^{N},
                \alpha,\varepsilon_{i,k|t},p}$
            \EndFor

            \State
            $\bar{\rho}_{k|t}^{\mathrm{DR},m}
            \gets
            \max_{i=1,\ldots,n_o}
            \bar{\rho}_{i,k|t}^{\mathrm{DR},m}$
        \EndFor

        \State
        $\widetilde{S}^m
        \gets
        S^m+
        \beta\sum_{k=0}^{K-1}
        \left[
        \bar{\rho}_{k|t}^{\mathrm{DR},m}
        -\bar{\rho}
        \right]^+$
    \EndFor

    \State Compute $\{w^m\}_{m=1}^{M}$ from
    \eqref{eqn:mppi_weights} using $\widetilde{S}^m$
    \State
    $u_k\gets\sum_{m=1}^{M}w^m v_k^m$,
    \quad $k=0,\ldots,K-1$
    \State Apply $u(t)\gets u_0$
    \State
    $\mathbf{u}\gets
    (u_1,\ldots,u_{K-1},u_{K-1})$
\EndWhile
\end{algorithmic}
\end{algorithm}

\section{Experiments}
\label{sec:experiments}

We evaluate the proposed DR$^2$-MPPI algorithm in a simulated ground-vehicle roundabout, a simulated Crazyflie scenario, and physical Crazyflie experiments. The evaluation considers collision avoidance, task performance, and computation time under finite-sample obstacle predictions and prediction-model mismatch.

\subsection{Evaluation Protocol}

All controllers are implemented in Python using PyTorch. We compare standard MPPI~\cite{williams2018information}, RA-MPPI~\cite{yin2023risk}, DRA-MPPI~\cite{trevisan2025dynamic}, and DR$^2$-MPPI. Within each scenario, the controllers share the system dynamics, nominal task cost, environment, and core MPPI parameters; they differ only in their obstacle-risk evaluation and associated penalty.\footnote{Simulations were executed on a workstation with an Intel Core i9-14900K CPU, an NVIDIA GeForce RTX 5080 GPU, and 128~GB of RAM.}

\textbf{Metrics:}
A collision occurs when the occupied regions of the ego robot and an obstacle intersect. The run-minimum signed clearance is the smallest signed clearance attained during an execution. A run is successful if the prescribed task is completed without collision. Collision counts, clearances, and planning times include all runs, whereas task costs include successful runs only. Controller-specific risk penalties are excluded from the reported task cost. Planning time is measured per update using a wall-clock timer, with CUDA synchronization before and after each measurement. We use $d_s=0$ unless otherwise stated, so zero signed clearance corresponds to physical contact.

\begin{algorithm}[t!]
\caption{Empirical DR-CVaR Upper-Bound Evaluation}
\label{alg:drcvar_eval}
\begin{algorithmic}[1]
\Function{EvalDRCVaR}{$\{L^j\}_{j=1}^{N},\alpha,\eta,p$}
    \State Sort
    $L^{(1)}\leq\cdots\leq L^{(N)}$
    \State $q\gets\lceil\alpha N\rceil$
    \State
    $\displaystyle
    \hat{\rho}\gets
    \frac{
    \sum_{j=q+1}^{N}L^{(j)}
    +(q-\alpha N)L^{(q)}
    }{N(1-\alpha)}$
    \State
    $\displaystyle
    \hat{\rho}^{\mathrm{DR}}
    \gets
    \hat{\rho}
    +
    \frac{\eta}{(1-\alpha)^{1/p}}$
    \State \Return $\hat{\rho}^{\mathrm{DR}}$
\EndFunction
\end{algorithmic}
\end{algorithm}

\textbf{Ablation design:}
We compare four CVaR-based configurations. \emph{Nominal RA-MPPI} follows~\cite{yin2023risk} with $d_s=0$. The \emph{DR$^2$-MPPI architecture ablation} sets $\varepsilon=0$, preserving the proposed stage-wise aggregation and penalty while removing the Wasserstein correction. \emph{Margin-matched RA-MPPI} retains the nominal architecture and sets $d_s=d_s^{\mathrm{eff}}=\varepsilon^\star/(1-\alpha)^{1/p}$. Finally, \emph{DR$^2$-MPPI} uses $d_s=0$ and $\varepsilon=\varepsilon^\star$. Comparing nominal RA-MPPI with the $\varepsilon=0$ ablation isolates the architectural change; comparing DR$^2$-MPPI with $\varepsilon=0$ and $\varepsilon=\varepsilon^\star$ isolates the robustness correction; and comparing margin-matched RA-MPPI with DR$^2$-MPPI at $\varepsilon^\star$ compares architectures at equal effective clearance. Hereafter, unqualified DR$^2$-MPPI denotes the nonzero-radius configuration.

\textbf{Radius selection:}
For each simulation scenario, we first sweep a common radius $\varepsilon_{i,k|t}\equiv\varepsilon$ in preliminary runs. We then select a radius that provides an appropriate observed safety--performance tradeoff and hold it fixed in a separate 100-run controller comparison. This is an empirical tuning procedure rather than a statistical calibration of ambiguity-set coverage. Accordingly, the experiments assess performance under the encountered prediction mismatch but do not provide a probabilistic safety guarantee.

\subsection{Ground-Vehicle Simulation}

\subsubsection{Setup}

A differential-drive vehicle follows a reference path through a single-lane roundabout while surrounding vehicles follow prerecorded trajectories that create potential conflicts. All vehicles have rectangular footprints of length $0.8\,\mathrm{m}$ and width $0.3\,\mathrm{m}$. The ego state is $x_k=[p_{x,k},p_{y,k},\psi_k]^\top$, the control is $v_k=[\nu_k,\omega_k]^\top$, and the sampling period is $\Delta t=0.04\,\mathrm{s}$. The standard discrete unicycle model is used.

At each obstacle-observation update, the measured position, heading, speed, longitudinal acceleration, and yaw rate are propagated using an approximate kinematic model. Observations and predictions are refreshed every 10 control steps. Between updates, surrounding vehicles continue to move while the planner reuses the previous predictions, creating a mismatch between predicted and realized motion.

For obstacle $i$ and prediction step $k$, the predictor generates
\[
\theta_{i,t+k|t}^{j}
=
\left(
\bar p_{i,t+k|t}+\epsilon_{i,k}^{j},
\bar R_{i,t+k|t}
\right), \; \epsilon_{i,k}^{j}
\sim
\mathcal N\!\left(
0,\operatorname{diag}(0.07^2,0.07^2)
\right), \quad j=1,\ldots,N.
\]
Thus, position is sampled while the predicted orientation is shared across samples. Finite sampling approximates the reference distribution, while the approximate predictor and intermittent observations produce model mismatch.

We use $M=1200$ control rollouts, $K=20$, $\lambda=2.0$, and $N=10$. The $0.8\,\mathrm{s}$ prediction horizon uses $\nu_k\in[0,5]\,\mathrm{m/s}$, $\omega_k\in[-2.5,2.5]\,\mathrm{rad/s}$, and reference speed $\nu_{\mathrm{des}}=2.8\,\mathrm{m/s}$. Control perturbations have covariance
$\Sigma=\operatorname{diag}(2.0^2,1.04^2)$.
DRA-MPPI uses $N_{\mathrm{mc}}=10$, $\sigma_{\mathrm{cp}}=0$, $\omega_{\mathrm{soft}}=100$, and $\omega_{\mathrm{hard}}=10$. The CVaR controllers use $\alpha=0.95$, $\beta=100$, and $\bar\rho=0$.

\begin{figure}[t]
    \centering
    \includegraphics[width=0.6\linewidth]{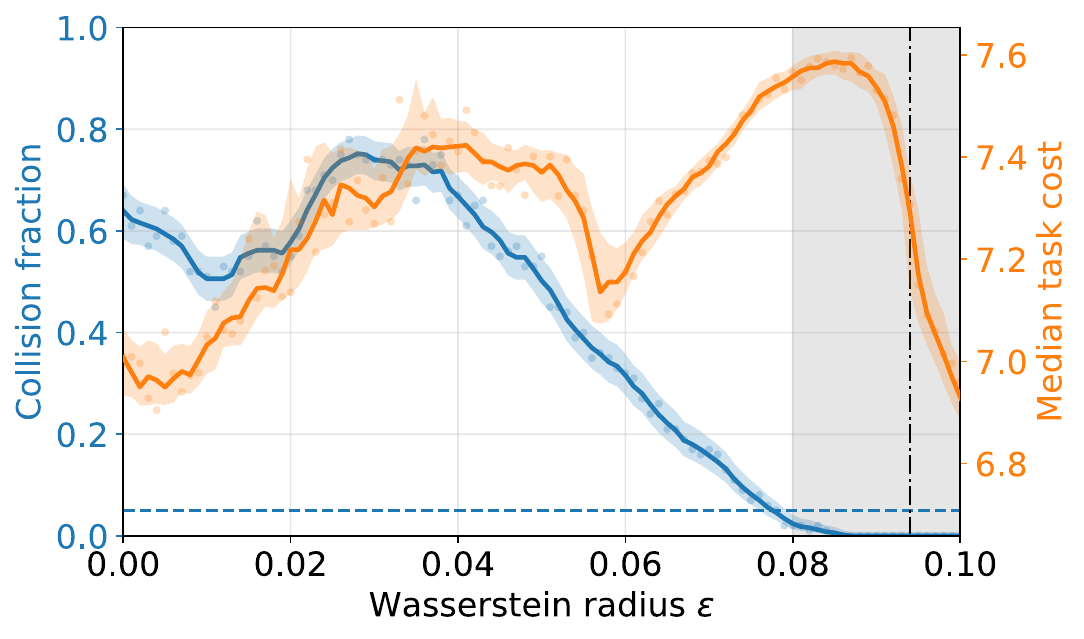}
    \caption{
    Ground-vehicle radius sweep (100 runs/radius). Smoothed collision fraction (blue) and median task cost (orange) are shown with 95\% confidence bands. The gray region begins at the smallest tested $\varepsilon$ whose upper Wilson bound is below 0.05; dashed and dash-dotted lines mark the 0.05 threshold and $\varepsilon^\star=0.094$, respectively.
    }
    \label{fig:dr_epsilon}
\end{figure}

\begin{table}[t]
\centering
\caption{Simulation results over 100 runs per controller and scenario.
Continuous metrics are reported as mean $\pm$ SD. Clearances and
planning times include all runs; costs include successful runs only.
The risk setting specifies the prescribed clearance $d_s$ or
Wasserstein radius $\varepsilon$.}
\label{tab:simulation_results}

\footnotesize
\renewcommand{\arraystretch}{1.14}

\begin{tabular}{@{}clccccc@{}}
\toprule
&
\textbf{Method}
&
\shortstack{\textbf{Risk}\\\textbf{setting (m)}}
&
\shortstack{\textbf{Collision}\\\textbf{runs}\\$n/100$}
&
\shortstack{\textbf{Run-min. signed}\\\textbf{clearance (m)}}
&
\shortstack{\textbf{Successful-run}\\\textbf{cost}}
&
\shortstack{\textbf{Planning}\\\textbf{time (ms)}}
\\
\midrule

\multirow{6}{*}{
    \rotatebox[origin=c]{90}{\textbf{Ground vehicle}}
}
& MPPI
& --
& 100
& $-0.158 \pm 0.049$
& --
& $18.85 \pm 2.59$
\\
\cmidrule(lr){2-7}

& DRA-MPPI
& --
& 24
& $0.079 \pm 0.097$
& $6.07 \pm 0.30$
& $65.16 \pm 1.98$
\\
\cmidrule(lr){2-7}

& \multirow{2}{*}{RA-MPPI}
& $d_s=0$
& 52
& $-0.033 \pm 0.103$
& $6.66 \pm 0.13$
& $18.79 \pm 2.10$
\\

&
& $d_s=1.88$
& 0
& $1.980 \pm 0.321$
& $9.95 \pm 4.41$
& $19.27 \pm 3.20$
\\
\cmidrule(lr){2-7}

& \multirow{2}{*}{DR$^2$-MPPI}
& $\varepsilon=0$
& 67
& $-0.046 \pm 0.091$
& $6.81 \pm 0.17$
& $18.73 \pm 1.35$
\\

&
& $\varepsilon=0.094$
& 0
& $0.765 \pm 0.451$
& $7.29 \pm 0.50$
& $19.20 \pm 3.05$
\\

\midrule

\multirow{6}{*}{
    \rotatebox[origin=c]{90}{\textbf{Crazyflie}}
}
& MPPI
& --
& 100
& $-0.029 \pm 0.007$
& --
& $11.53 \pm 0.34$
\\
\cmidrule(lr){2-7}

& DRA-MPPI
& --
& 95
& $-0.010 \pm 0.008$
& $1189.59 \pm 30.51$
& $14.13 \pm 0.43$
\\
\cmidrule(lr){2-7}

& \multirow{2}{*}{RA-MPPI}
& $d_s=0$
& 47
& $0.000 \pm 0.005$
& $1301.12 \pm 58.89$
& $11.36 \pm 0.34$
\\

&
& $d_s=0.03$
& 2
& $0.013 \pm 0.006$
& $2044.13 \pm 119.88$
& $11.39 \pm 0.38$
\\
\cmidrule(lr){2-7}

& \multirow{2}{*}{DR$^2$-MPPI}
& $\varepsilon=0$
& 37
& $0.002 \pm 0.005$
& $1409.65 \pm 68.72$
& $11.58 \pm 0.35$
\\

&
& $\varepsilon=0.0015$
& 1
& $0.010 \pm 0.005$
& $1720.76 \pm 89.40$
& $11.68 \pm 1.04$
\\

\bottomrule
\end{tabular}
\end{table}

\subsubsection{Results}

Figure~\ref{fig:dr_epsilon} shows that the observed collision frequency is nonmonotonic at small radii but generally decreases beyond $\varepsilon\approx0.04$ and falls below $0.05$ near $\varepsilon=0.08$. We select $\varepsilon^\star=0.094$, which yields $d_s^{\mathrm{eff}}
=
1.88\,\mathrm{m}$.
The margin-matched RA-MPPI controller therefore uses $d_s=1.88\,\mathrm{m}$.

The aggregate results in Table~\ref{tab:simulation_results} show that nominal RA-MPPI records 52 collision runs, while DR$^2$-MPPI with $\varepsilon=0$ records 67. Thus, the architectural change alone does not improve safety in this scenario; the nonzero correction is essential. Both margin-matched RA-MPPI and DR$^2$-MPPI with $\varepsilon=\varepsilon^\star$ complete all 100 runs without collision. DR$^2$-MPPI nevertheless obtains lower successful-run cost than margin-matched RA-MPPI, $7.29$ versus $9.95$, while maintaining positive mean run-minimum clearance. Under the matched effective clearance, the proposed architecture therefore produces less conservative task behavior with the same observed collision count.

\begin{figure}[t]
    \centering
    {\small
    \legendbox{mppiblue}{MPPI}\hspace{0.4em}
    \legendbox{dramppiorange}{DRA-MPPI}\hspace{0.4em}
    \legendbox{ramppipurple}{RA-MPPI}\hspace{0.4em}
    \legendbox{drmppigreen}{DR$^2$-MPPI}\hspace{0.4em}
    \legendstar{Collision}\par}
    \vspace{0.5em}

    \begin{subfigure}[t]{0.4\columnwidth}
        \includegraphics[width=\linewidth]{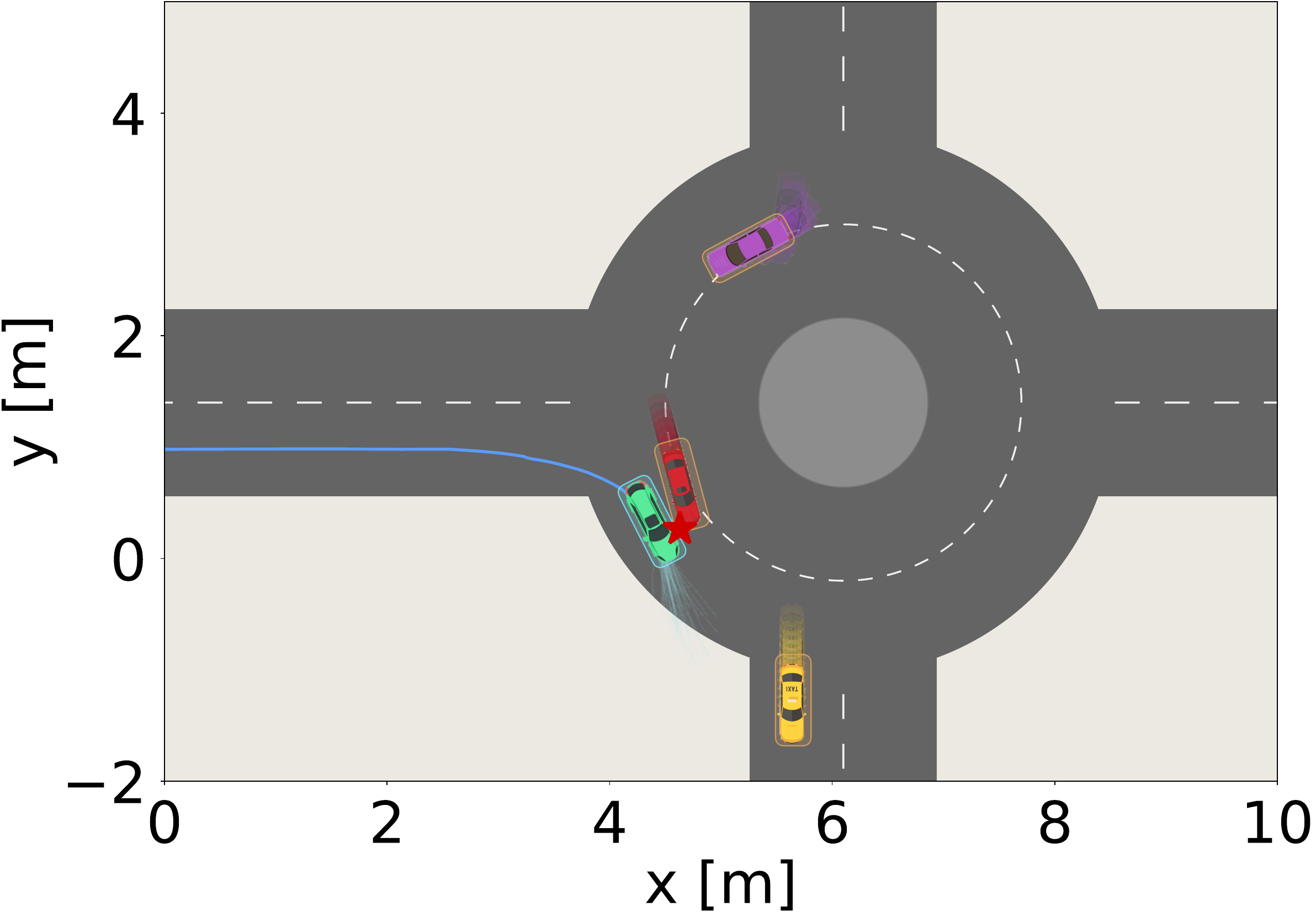}
    \end{subfigure}
    \begin{subfigure}[t]{0.4\columnwidth}
        \includegraphics[width=\linewidth]{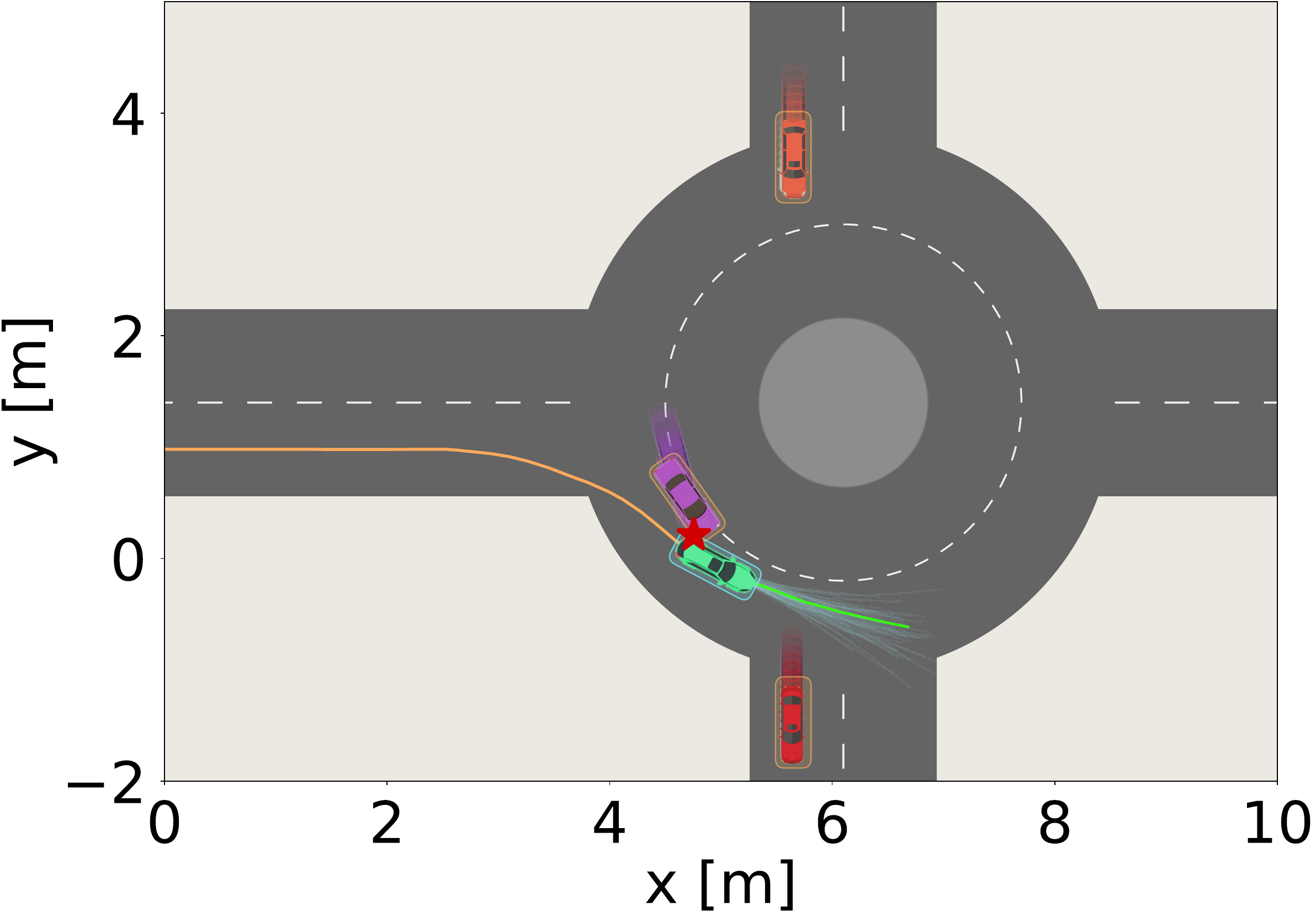}
    \end{subfigure}\\
    
    \vspace{0.3em}
    \begin{subfigure}[t]{0.4\columnwidth}
        \includegraphics[width=\linewidth]{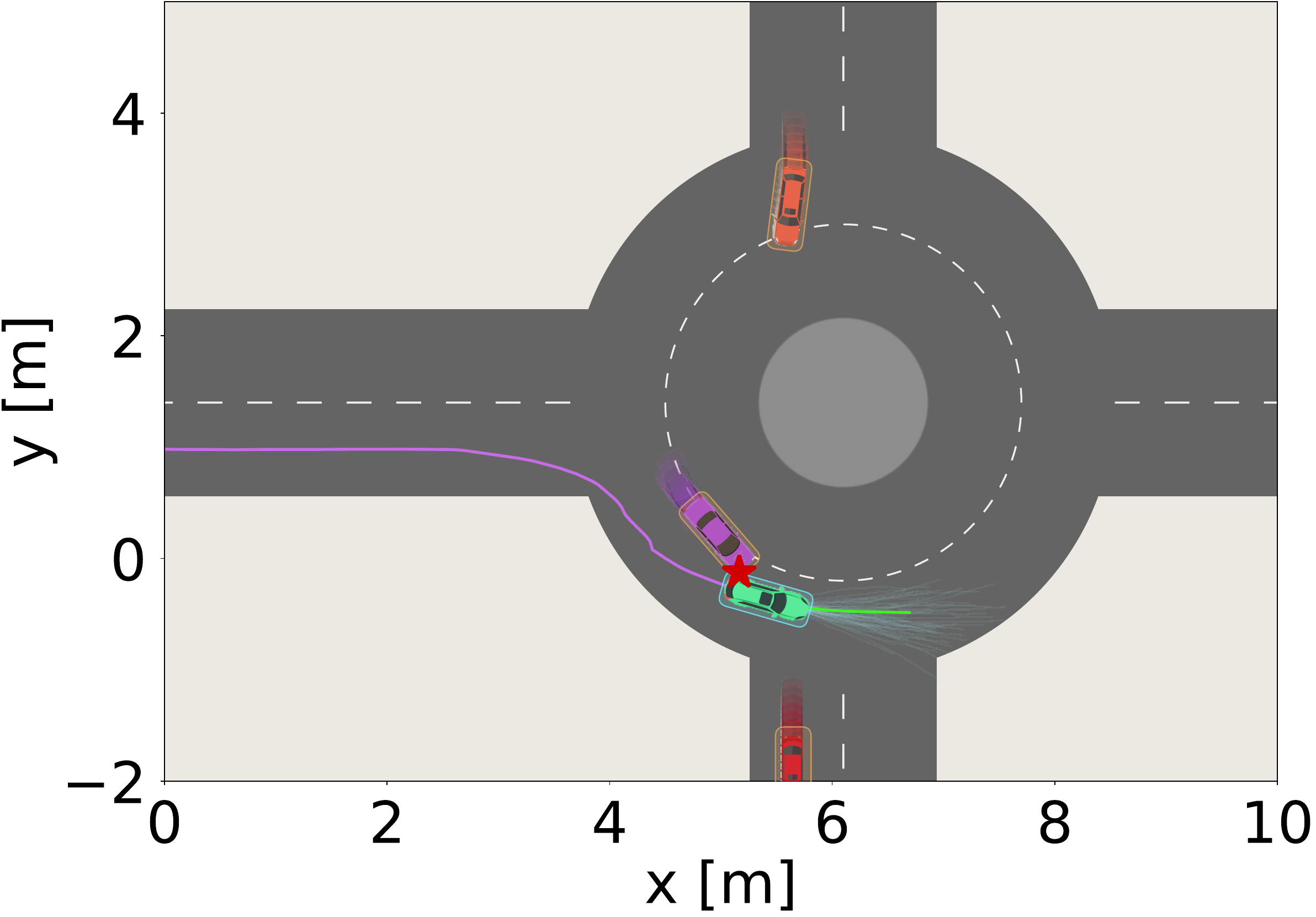}
    \end{subfigure}
    \begin{subfigure}[t]{0.4\columnwidth}
        \includegraphics[width=\linewidth]{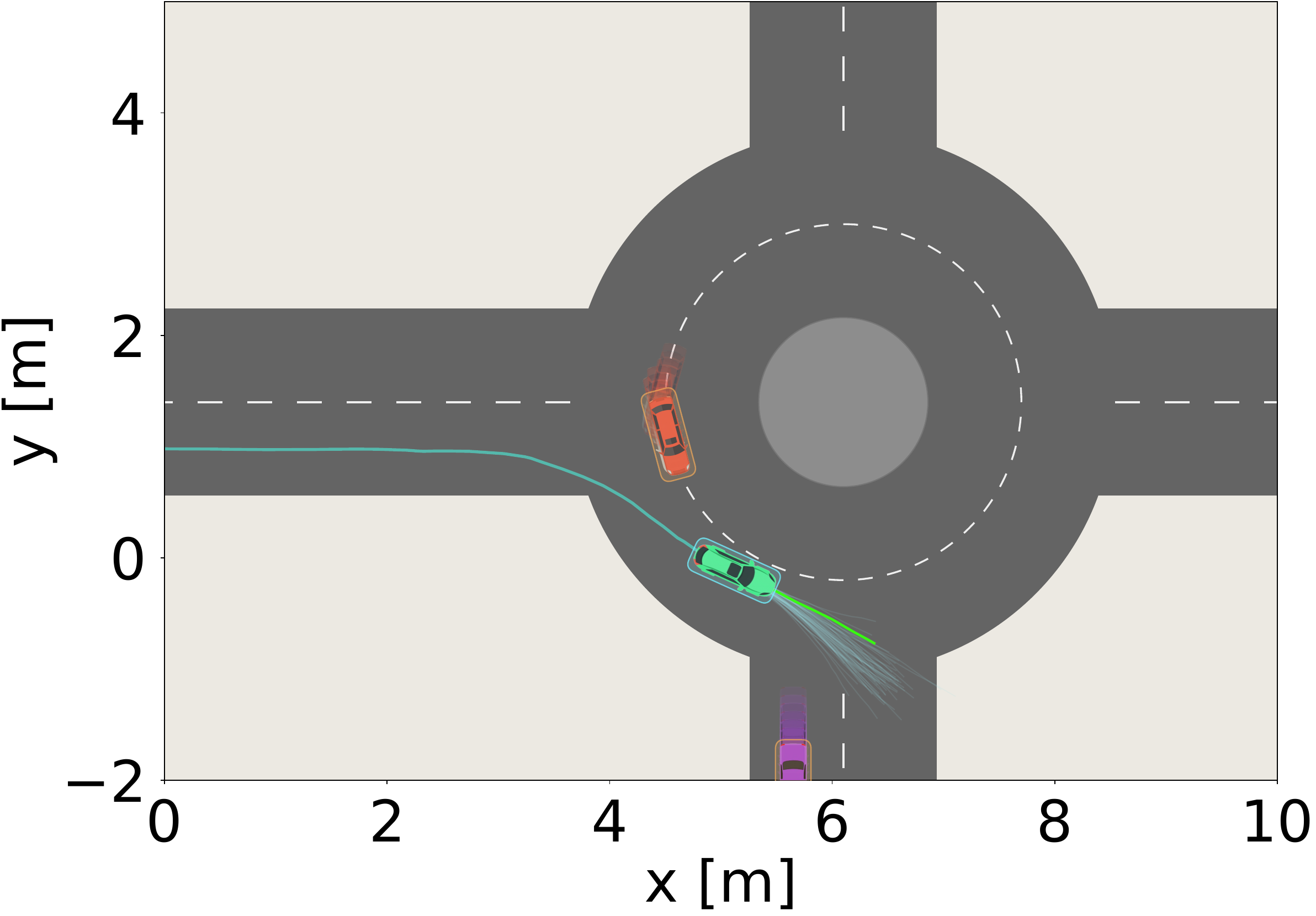}
    \end{subfigure}

    \caption{Representative ground-vehicle executions. Red stars mark collision locations. DR$^2$-MPPI yields before entering the conflict region and proceeds after the surrounding traffic clears.}
    \label{fig:planner_comparison}
\end{figure}

Figure~\ref{fig:planner_comparison} illustrates representative executions. MPPI and DRA-MPPI enter the conflict region before the traffic clears, whereas DR$^2$-MPPI yields and proceeds when a safer gap becomes available.

\subsection{Crazyflie Simulation}

\subsubsection{Setup}

An ego Crazyflie tracks a three-dimensional figure-eight trajectory among two moving Crazyflie obstacles. Each vehicle is modeled as a sphere of radius $0.04\,\mathrm{m}$. The rollout state is
$x_k=[p_k^\top,v_k^\top,\phi_k,\vartheta_k,\psi_k]^\top$,
and the control contains the commanded roll, pitch, yaw, and collective thrust. CrazySim~\cite{llanes2024crazysim} is used with $\Delta t=0.02\,\mathrm{s}$.

The obstacles follow figure-eight trajectories with time-varying velocities and a temporal offset. Predictions are refreshed every 15 control steps, or $0.3\,\mathrm{s}$, and reused between observations. Pose samples are generated as 
\[
\theta_{i,t+k|t}^{j}
=
\left(
\bar p_{i,t+k|t}+\epsilon_{i,k}^{j},
\bar R_{i,t+k|t}
\right),\;
\epsilon_{i,k}^{j}
\sim
\mathcal N\!\left(
0,\operatorname{diag}(0.01^2,0.01^2,0.01^2)
\right), \quad j=1,\dots, N.
\]
Mismatch arises because predictions are reused while the obstacles continue along trajectories with changing velocities.

All controllers use $M=2000$, $K=25$, $\lambda=5.0$, and a $0.5\,\mathrm{s}$ horizon. Control perturbations have covariance
$\operatorname{diag}(0.1^2,0.1^2,0.01^2,0.08^2)$.
Commanded attitudes are limited to $\pm20^\circ$, and thrust is constrained to $[0,0.4776]\,\mathrm{N}$. DRA-MPPI uses $N_{\mathrm{mc}}=10$, $\sigma_{\mathrm{cp}}=0.05$, $\omega_{\mathrm{soft}}=3000$, and $\omega_{\mathrm{hard}}=1000$. The CVaR controllers use $N=10$, $\alpha=0.95$, $\beta=2000$, and $\bar\rho=0$.
The radius sweep selects $\varepsilon^\star=0.0015$, corresponding to
$d_s^{\mathrm{eff}}=0.03\,\mathrm{m}$. Margin-matched RA-MPPI therefore uses $d_s=0.03\,\mathrm{m}$.

\subsubsection{Results}

\begin{figure*}[!t]
    \centering
    \small
    \legendbox{mppiblue}{MPPI}\hspace{1.0em}
    \legendbox{dramppiorange}{DRA-MPPI}\hspace{1.0em}
    \legendbox{ramppipurple}{RA-MPPI}\hspace{1.0em}
    \legendbox{drmppigreen}{DR$^2$-MPPI}\hspace{1.0em}
    \legendstar{Collision}
    \par\vspace{0.4em}

    \begin{subfigure}[b]{0.45\textwidth}
    \centering
        \includegraphics[width=\linewidth]{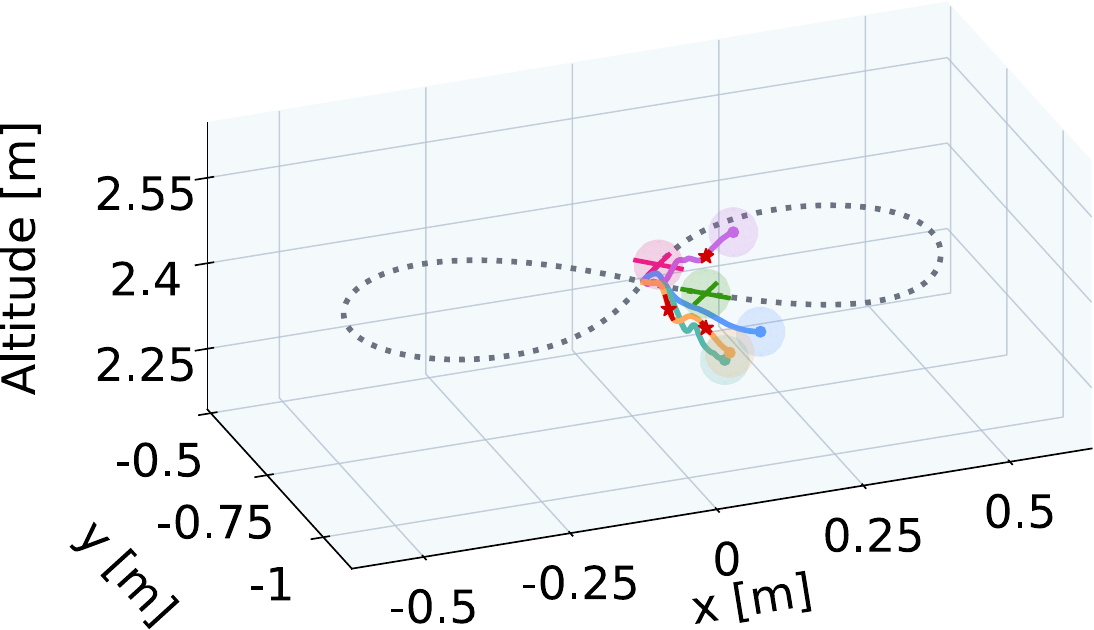}
        \caption{Step 52}
    \end{subfigure}
    \hfill
    \begin{subfigure}[b]{0.45\textwidth}
    \centering
        \includegraphics[width=\linewidth]{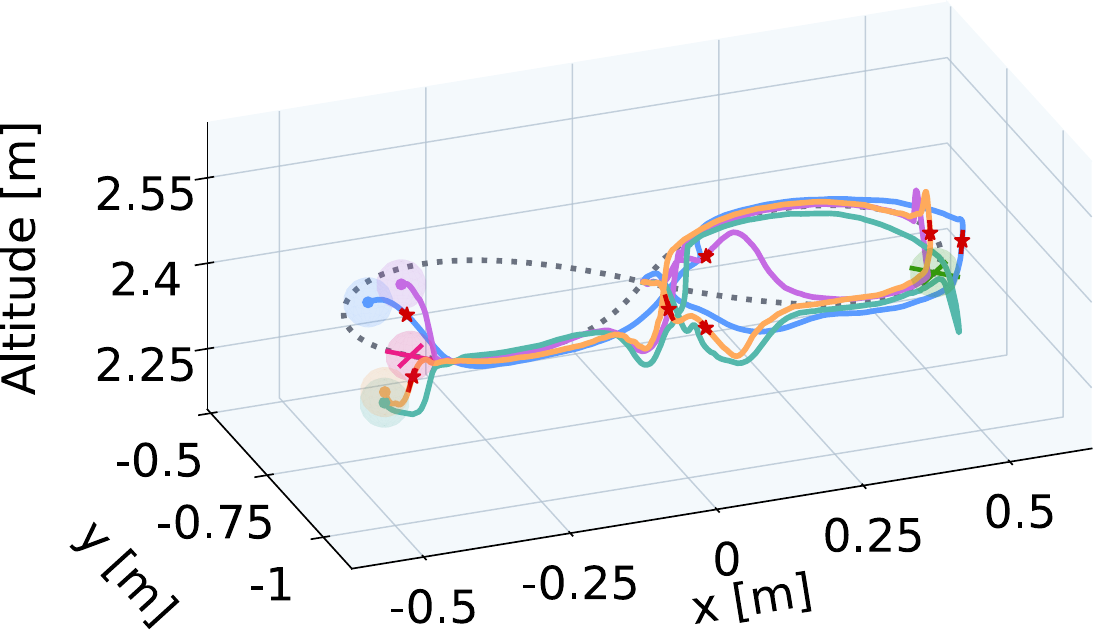}
        \caption{Step 344}
    \end{subfigure}\\
    \begin{subfigure}[b]{0.45\textwidth}
    \centering
        \includegraphics[width=\linewidth]{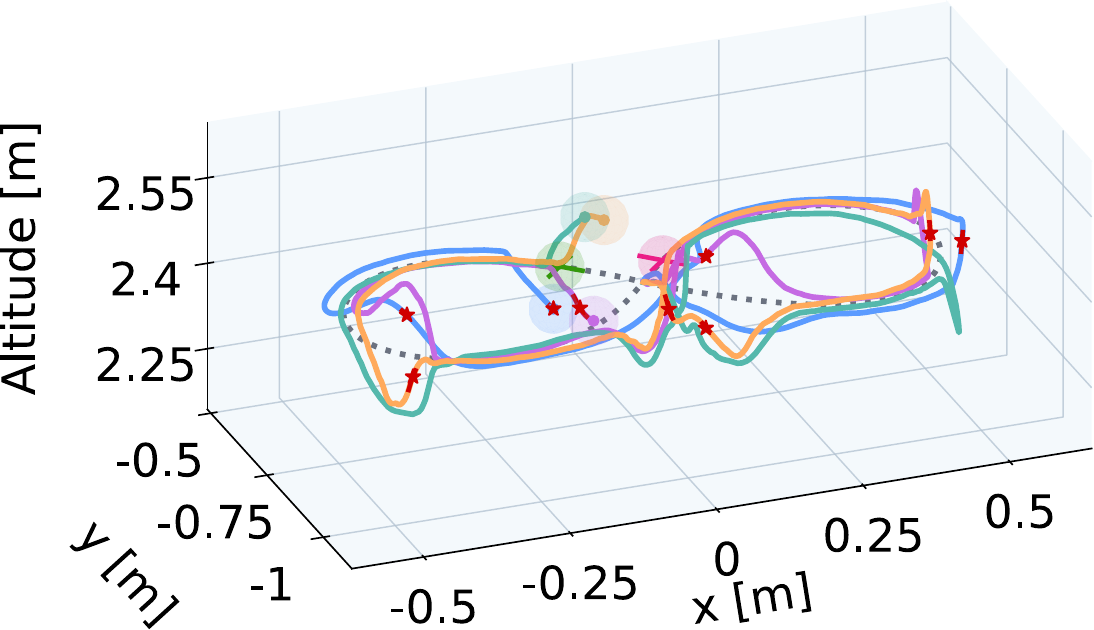}
        \caption{Step 456}
    \end{subfigure}

    \caption{Representative Crazyflie simulation trajectories. Dotted curves show reference paths, and red stars mark collisions. DR$^2$-MPPI begins avoidance earlier, maintains greater separation, and subsequently returns toward the reference path.}
    \label{fig:crazyflie_comparison}
\end{figure*}

\Cref{tab:simulation_results} shows that MPPI and DRA-MPPI collide in 100 and 95 runs, respectively. Nominal RA-MPPI records 47 collision runs, whereas the $\varepsilon=0$ DR$^2$-MPPI ablation records 37, suggesting a benefit from the proposed architecture in this scenario. At matched effective clearance, the collision counts fall to two for RA-MPPI and one for DR$^2$-MPPI, while DR$^2$-MPPI achieves lower successful-run cost. \Cref{fig:crazyflie_comparison} illustrates the corresponding behavior: DR$^2$-MPPI initiates avoidance earlier, deviates farther from the reference, and returns toward it after the conflict.

\begin{figure*}[!t]
    \centering
    \small
    \legendbox{mppiblue}{MPPI}\hspace{1.0em}
    \legendbox{dramppiorange}{DRA-MPPI}\hspace{1.0em}
    \legendbox{ramppipurple}{RA-MPPI}\hspace{1.0em}
    \legendbox{drmppigreen}{DR$^2$-MPPI}\hspace{1.0em}
    \legendstar{Collision}
    \par\vspace{0.4em}

    \begin{subfigure}[t]{0.49\textwidth}
        \includegraphics[width=\linewidth]{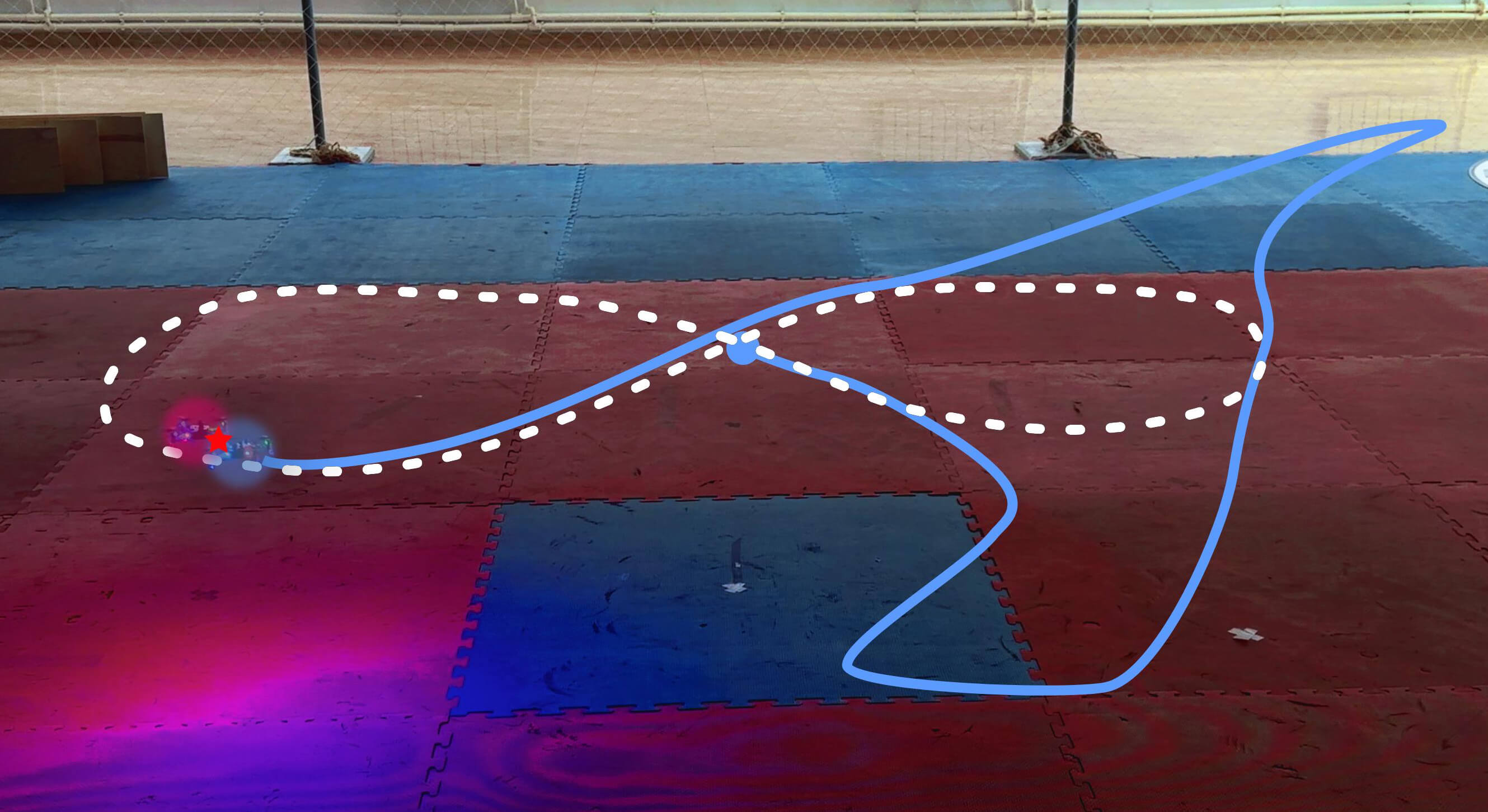}
    \end{subfigure}
    \begin{subfigure}[t]{0.49\textwidth}
        \includegraphics[width=\linewidth]{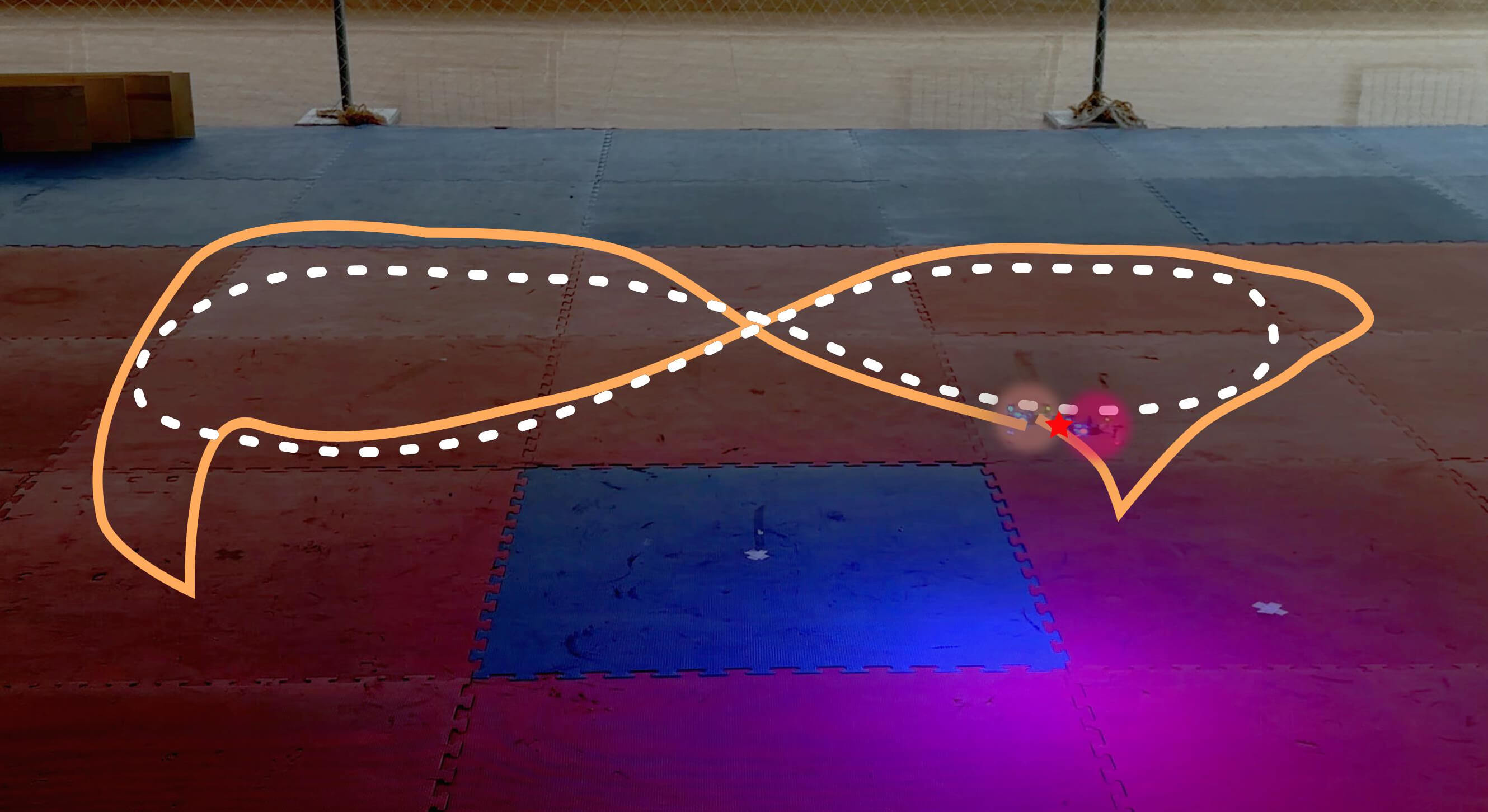}
    \end{subfigure}\\\vspace{0.3em}
    \begin{subfigure}[t]{0.49\textwidth}
        \includegraphics[width=\linewidth]{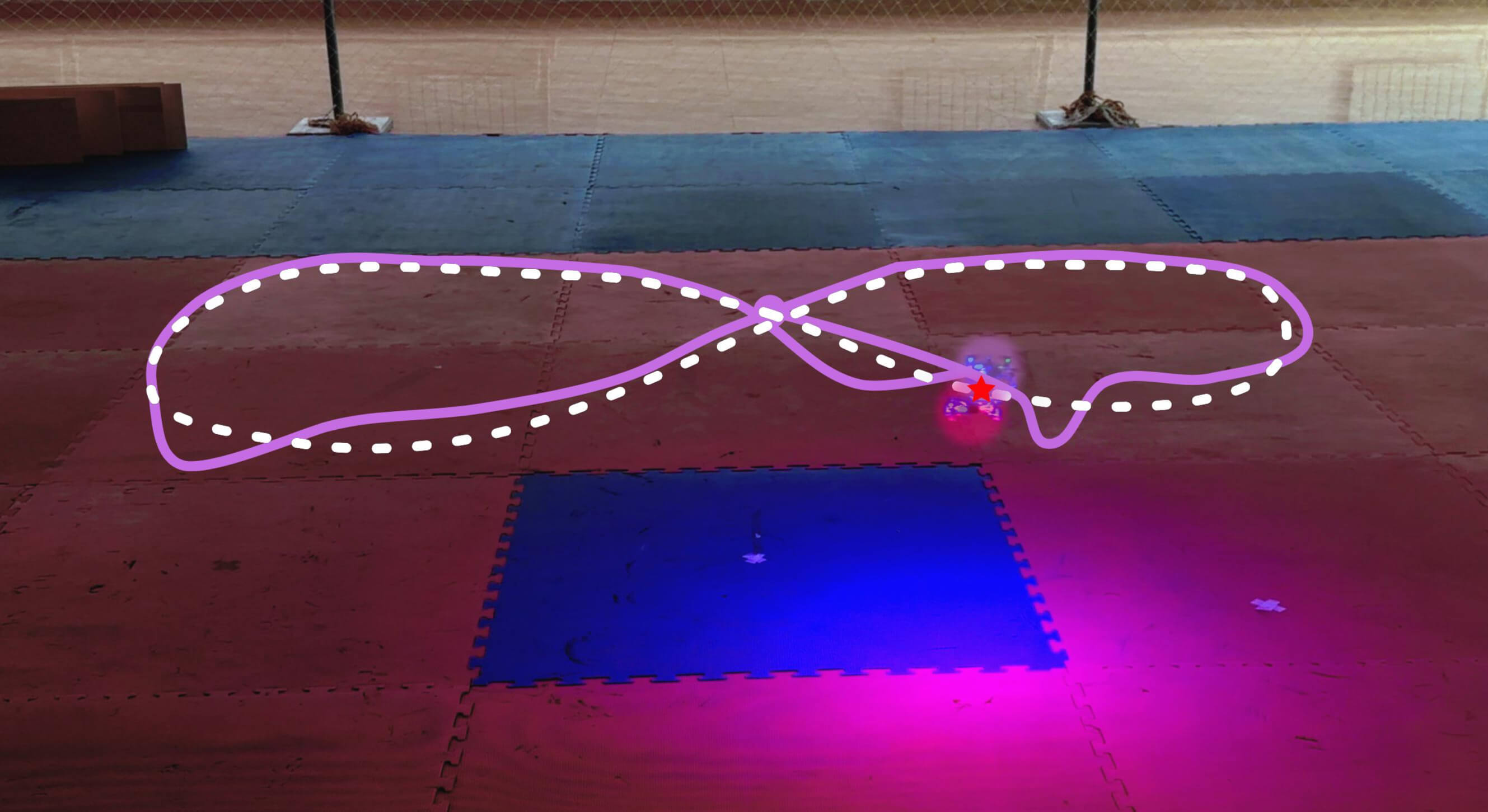}
    \end{subfigure}
    \begin{subfigure}[t]{0.49\textwidth}
        \includegraphics[width=\linewidth]{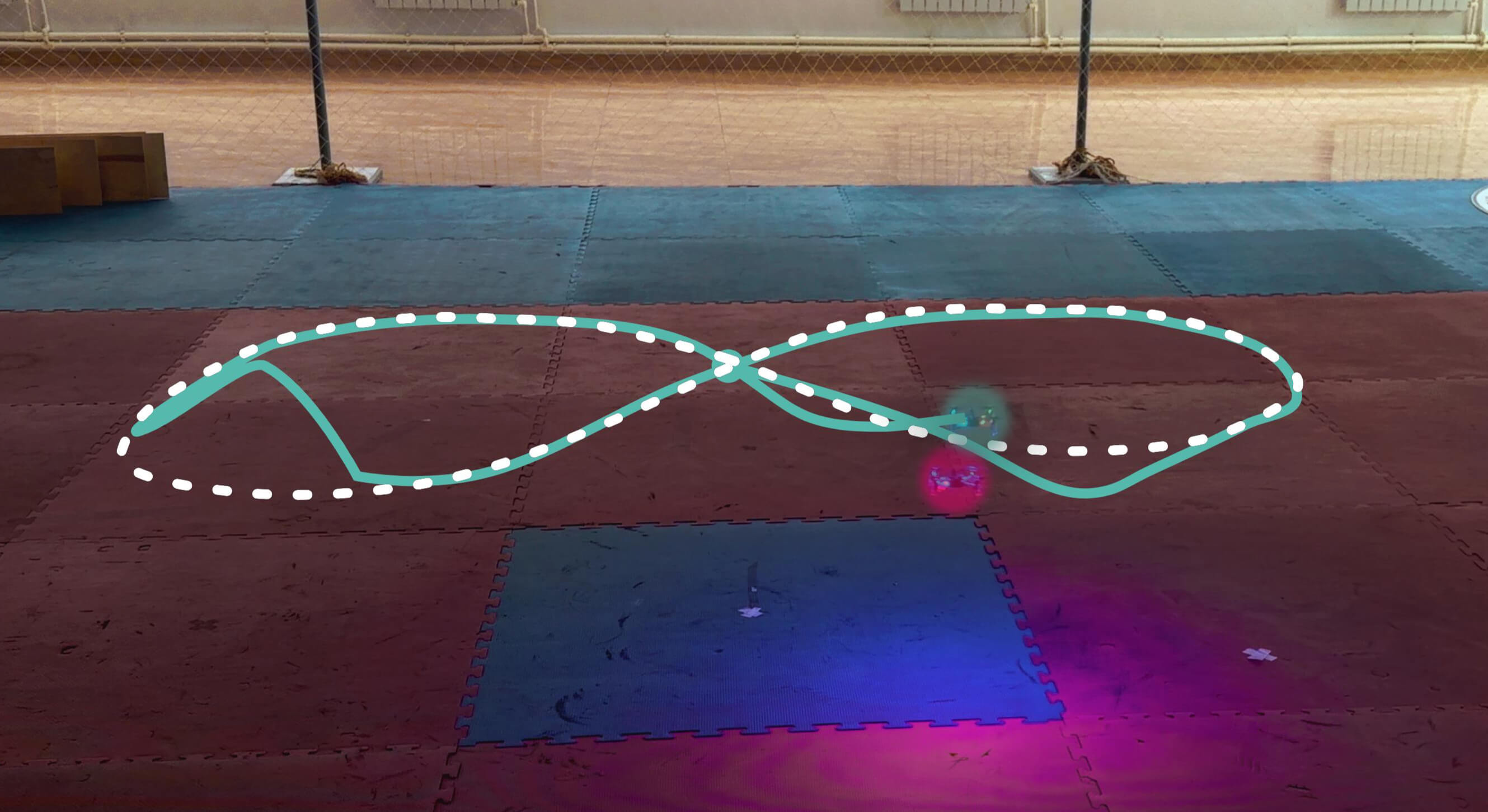}
    \end{subfigure}

    \caption{Representative hardware runs for MPPI (top left), DRA-MPPI (top right), RA-MPPI (bottom left), and DR$^2$-MPPI (bottom right). The latter three use the 10-sample configuration. Dashed curves show reference paths, and red stars mark collisions. In the run shown, DR$^2$-MPPI completes two laps with four obstacle encounters and no contact. A video of the experiments is available at \url{https://youtu.be/l1w6ZpRgT2I}.}
    \label{fig:crazyflie_hardware}
\end{figure*}

\subsection{Crazyflie Hardware Experiments}

\subsubsection{Setup}

Two Crazyflie quadrotors operate in an indoor motion-capture environment. The ego vehicle tracks a Poly4D figure-eight trajectory, while a second Crazyflie follows the same trajectory in the opposite direction as a moving obstacle. Each run comprises two laps and four encounters. Five runs are performed for each controller configuration.

Motion-capture measurements provide ego and obstacle positions at $50\,\mathrm{Hz}$. Obstacle velocity is estimated by finite differences followed by low-pass filtering, and a constant-velocity model predicts its position over the planning horizon. Gaussian perturbations with covariance
$\operatorname{diag}(0.01^2,0.01^2,0.01^2)$
are added to the nominal predictions. Unless otherwise stated, the geometry, rollout model, horizon, costs, and control limits match the Crazyflie simulation.

The reference predictions do not include acceleration, estimation and filtering errors, or aerodynamic coupling between nearby vehicles. During some encounters, the obstacle exhibits transient motion disturbances as the ego vehicle passes above it. These unmodeled effects create deviations between predicted and realized obstacle poses.

We evaluate DRA-MPPI with \(N_{\mathrm{mc}}\in\{10,50\}\), RA-MPPI with \(N\in\{10,50\}\), and DR²-MPPI with \(N=10\) and \(\epsilon\in\{0,0.0085\}\); MPPI has no prediction-sample-count setting.
All CVaR configurations use $\alpha=0.95$. The selected radius $\varepsilon^\star=0.0085$ corresponds to $d_s^{\mathrm{eff}}=0.17\,\mathrm{m}$.

\subsubsection{Results}

Table~\ref{tab:hardware_results} reports the five-run results. MPPI and the 10-sample RA-MPPI configuration collide in four of five runs, while 10-sample DRA-MPPI collides in all five. Increasing the sample count does not produce collision-free completion for either baseline. DR$^2$-MPPI with $\varepsilon=0$ collides in three runs, whereas DR$^2$-MPPI with $\varepsilon=\varepsilon^\star$ completes all five runs without collision and attains a mean run-minimum signed clearance of $0.077\,\mathrm{m}$.

Figure~\ref{fig:crazyflie_hardware} shows representative 10-sample executions. DR$^2$-MPPI initiates avoidance before the conflict region and completes both laps and all four encounters without physical contact. The five-run results show repeatable hardware feasibility under the encountered mismatch, not a statistical safety guarantee.

\begin{table}[t]
\centering
\caption{Hardware results over five runs per controller setting.
Continuous metrics are reported as mean $\pm$ SD. Clearances and
planning times include all runs; costs include successful runs only.
Singleton costs are reported without an SD.}
\label{tab:hardware_results}

\footnotesize
\renewcommand{\arraystretch}{1.15}

\begin{tabular}{@{}lcccccc@{}}
\toprule
\textbf{Method}
&
\shortstack{\textbf{Risk}\\\textbf{setting (m)}}
&
\shortstack{\textbf{Sample}\\\textbf{Size}}
&
\shortstack{\textbf{Collision}\\\textbf{runs}\\$n/5$}
&
\shortstack{\textbf{Run-min. signed}\\\textbf{clearance (m)}}
&
\shortstack{\textbf{Successful-run}\\\textbf{cost} ($\times10^4$)}
&
\shortstack{\textbf{Planning}\\\textbf{time (ms)}}
\\
\midrule

MPPI
& --
& --
& 4
& $-0.017 \pm 0.024$
& $32.325$
& $16.944 \pm 1.661$
\\
\cmidrule(lr){1-7}

\multirow{2}{*}{DRA-MPPI}
& --
& 10
& 5
& $-0.046 \pm 0.025$
& --
& $23.305 \pm 0.966$
\\

& --
& 50
& 3
& $0.062 \pm 0.195$
& $32.030 \pm 44.760$
& $20.750 \pm 3.569$
\\
\cmidrule(lr){1-7}

\multirow{2}{*}{RA-MPPI}
& $d_s=0$
& 10
& 4
& $-0.004 \pm 0.017$
& $1.854$
& $16.858 \pm 1.572$
\\

& $d_s=0$
& 50
& 4
& $-0.007 \pm 0.016$
& $14.458$
& $15.403 \pm 0.356$
\\
\cmidrule(lr){1-7}

\multirow{2}{*}{DR$^2$-MPPI}
& $\varepsilon=0$
& 10
& 3
& $0.018 \pm 0.019$
& $1.895 \pm 0.693$
& $15.618 \pm 0.315$
\\

& $\varepsilon=0.0085$
& 10
& 0
& $0.077 \pm 0.016$
& $2.571 \pm 0.514$
& $15.739 \pm 0.284$
\\

\bottomrule
\end{tabular}
\end{table}

\section{Conclusion}

This paper presented DR$^2$-MPPI for risk-aware control under ambiguity in dynamic-obstacle predictions. Using Wasserstein ambiguity sets and the $1$-Lipschitz signed-clearance loss, we derived a tractable worst-case CVaR bound consisting of empirical CVaR and an analytic correction. Its equivalent clearance-inflation interpretation incorporates distributional robustness without nested optimization or changes to MPPI's parallel sampling structure. Simulations and hardware experiments demonstrated improved observed separation under prediction mismatch and real-time feasibility, while ablations distinguished the architecture, robustness correction, and matched-clearance margin. Because the radii were selected empirically, these results do not constitute a safety guarantee. Future work will address statistical radius calibration, correlated predictions, and closed-loop guarantees.

\appendix

\section{Proof of Proposition~\ref{prop:drcvar_reform}}
\label{app:drcvar_reform}
\begin{proof}
Let $\pi\in\Pi(\mathbb{P},\mathbb{Q})$ be any coupling and let $(\xi,\xi')\sim\pi$. Define $X=g(\xi)$ and $Y=g(\xi')$. Since CVaR is law invariant, both random variables may be evaluated on the probability space induced by $\pi$.

Let $q$ be the Hölder conjugate of $p$. The dual representation of CVaR~\cite{shapiro2012minimax} gives
\begin{equation*}
\cvar_\alpha(X)
=
\sup_{\zeta\in\mathcal{Z}_\alpha}
\mathbb{E}_\pi[\zeta X],
\end{equation*}
where $\mathcal{Z}_\alpha
\coloneqq
\left\{
\zeta:
0\leq\zeta\leq\frac{1}{1-\alpha},
\ \mathbb{E}_\pi[\zeta]=1
\right\}$.
Consequently,
\begin{align*}
\left|\cvar_\alpha(X)-\cvar_\alpha(Y)\right|
&\leq
\sup_{\zeta\in\mathcal{Z}_\alpha}
\mathbb{E}_\pi[\zeta|X-Y|] \\
&\leq
\sup_{\zeta\in\mathcal{Z}_\alpha}
\|\zeta\|_q\,
\|X-Y\|_p.
\end{align*}
For every $\zeta\in\mathcal{Z}_\alpha$, 
$\|\zeta\|_q
\leq
(1-\alpha)^{-1/p}$,
with the usual interpretation $\|\zeta\|_\infty\leq(1-\alpha)^{-1}$ when $p=1$. Furthermore, the Lipschitz continuity of $g$ implies
\begin{equation*}
\|X-Y\|_p
\leq
\mathrm{Lip}(g)
\left(
\mathbb{E}_\pi[
d_\Xi(\xi,\xi')^p]
\right)^{1/p}.
\end{equation*}
Taking the infimum over $\pi\in\Pi(\mathbb{P},\mathbb{Q})$ proves~\eqref{eqn:cvar_stability}. Applying this inequality with $\mathbb{Q}=\hat{\mathbb{P}}$ and $W_p(\mathbb{P},\hat{\mathbb{P}})\leq\eta$ proves~\eqref{eqn:drcvar_ineq}.
\end{proof}

\section{Proof of~\Cref{lem:lipschitz_loss}}
\label{app:lipschitz_loss}

\begin{proof}
Fix the robot state $x$ and let $\theta=(p,R)$ and $\theta'=(p',R')$. For any $z\in\mathcal{O}_i^0$,
\begin{align*}
\left\|(Rz+p)-(R'z+p')\right\|_2
&\leq
\|p-p'\|_2+\|(R-R')z\|_2\\
&\leq
\|p-p'\|_2+
r_i\|R-R'\|_{\mathrm{op}},
\end{align*}
where $r_i=\sup_{z\in\mathcal{O}_i^0}\|z\|_2$. By matching points generated by the same $z$ in both directions, we obtain
\begin{equation*}
d_H\!\left(
\mathcal{O}_i(\theta),
\mathcal{O}_i(\theta')
\right)
\leq
\|p-p'\|_2+
r_i\|R-R'\|_{\mathrm{op}}.
\end{equation*}

For a compact obstacle set $\mathcal{O}$, define
\begin{equation*}
\gamma_{\mathcal{O}}(n)
\coloneqq
\inf_{o\in\mathcal{O}}n^\top o
-
\sup_{b\in\mathcal{B}(x)}n^\top b,
\qquad \|n\|_2=1.
\end{equation*}
By the signed-clearance definition, $\operatorname{sdist}\!\left(\mathcal{B}(x),\mathcal{O}\right)=\max_{\|n\|_2=1}\gamma_{\mathcal{O}}(n)$. For every unit vector $n$, the support-function Lipschitz property gives
\begin{equation*}
\left|
\gamma_{\mathcal{O}_i(\theta)}(n)
-
\gamma_{\mathcal{O}_i(\theta')}(n)
\right|
\leq
d_H\!\left(
\mathcal{O}_i(\theta),
\mathcal{O}_i(\theta')
\right).
\end{equation*}
Using $|\max f-\max g|\leq\max|f-g|$ and the definition of the loss, we obtain
\begin{align*}
\left|
\ell(x,\theta)-\ell(x,\theta')
\right|
&=
\left|
\operatorname{sdist}\!\left(
\mathcal{B}(x),\mathcal{O}_i(\theta)
\right)
-
\operatorname{sdist}\!\left(
\mathcal{B}(x),\mathcal{O}_i(\theta')
\right)
\right|\\
&\leq
d_H\!\left(
\mathcal{O}_i(\theta),
\mathcal{O}_i(\theta')
\right)\\
&\leq
\|p-p'\|_2+
r_i\|R-R'\|_{\mathrm{op}}\\
&=
d_{\Xi_i}(\theta,\theta'),
\end{align*}
which proves the claim.
\end{proof}

\bibliographystyle{IEEEtran}
\bibliography{ref}

\end{document}